\documentclass[3p,12pt,authoryear]{elsarticle}

\usepackage[utf8]{inputenc}
\usepackage[T1]{fontenc}
\usepackage[round]{natbib}
\usepackage{graphicx}
\usepackage{array}
\usepackage{caption}
\usepackage{float}
\usepackage{booktabs}
\usepackage{adjustbox}

\usepackage{amsmath,amssymb,amsfonts}
\usepackage{graphicx}
\usepackage{adjustbox}     
\usepackage{booktabs}      
\usepackage{array}         
\usepackage{caption}
\usepackage{subcaption}
\usepackage{float}
\usepackage{booktabs}
\usepackage{multirow}
\usepackage{adjustbox}
\usepackage{algorithm}
\usepackage{hyperref}
\usepackage{xcolor}
\usepackage{url}
\usepackage{amsmath}
\usepackage{adjustbox}
\usepackage{booktabs}
\usepackage{array}
\usepackage{float}
\usepackage{algorithm}
\usepackage{algpseudocode}
\usepackage{subcaption}
\usepackage{caption}
\definecolor{customgreen}{RGB}{0,170,85}

\usepackage{pdflscape}

\def\x{{\mathbf x}}

\usepackage{lineno}
\modulolinenumbers[5]

\biboptions{authoryear,sort&compress}

\begin{document}

\begin{frontmatter}

\title{Underwater Diffusion Attention Network \\with Contrastive Language-Image Joint Learning \\for Underwater Image Enhancement}
\vspace{2em}
\author{Afrah Shaahid}

\author{Muzammil Behzad\corref{cor1}}

\address{King Fahd University of Petroleum and Minerals, Saudi Arabia}
\ead{muzammil.behzad@kfupm.edu.sa}

\vspace{-0.5em}
\cortext[cor1]{Corresponding author}

\begin{abstract}
Underwater images are often affected by complex degradations such as light absorption, scattering, color casts, and artifacts, making enhancement critical for effective object detection, recognition, and scene understanding in aquatic environments. Existing methods, especially diffusion-based approaches, typically rely on synthetic paired datasets due to the scarcity of real underwater references, introducing bias and limiting generalization. Furthermore, fine-tuning these models can degrade learned priors, resulting in unrealistic enhancements due to domain shifts. To address these challenges, we propose UDAN-CLIP, an image-to-image diffusion framework pre-trained on synthetic underwater datasets and enhanced with a customized classifier based on vision-language model, a spatial attention module, and a novel CLIP-Diffusion loss. The classifier preserves natural in-air priors and semantically guides the diffusion process, while the spatial attention module focuses on correcting localized degradations such as haze and low contrast. The proposed CLIP-Diffusion loss further strengthens visual-textual alignment and helps maintain semantic consistency during enhancement. The proposed contributions empower our UDAN-CLIP model to perform more effective underwater image enhancement, producing results that are not only visually compelling but also more realistic and detail-preserving. These improvements are consistently validated through both quantitative metrics and qualitative visual comparisons, demonstrating the model’s ability to correct distortions and restore natural appearance in challenging underwater conditions.

    \end{abstract}
  \end{frontmatter}

\section{Introduction}
\label{sec:intro}

Ocean exploration has always been a promising domain of research for humanity. Approximately 71\% of the earth's surface is covered by oceans, which contain the majority of the earth's resources where the marine environment is acknowledged as an essential measure for achieving sustainable human developments \cite{a5}. This is because it offers important resources, including food, medicine, biology, and energy \cite{national1995understanding}. Despite its importance, most of the oceanic regions remain inaccessible or hazardous for humans to explore.

Underwater imaging plays a crucial role in diverse fields, including exploration and protection of marine ecology, coral reef monitoring, archaeology, underwater environmental surveillance, and underwater robotics \cite{a7, a8, a9, Raveendran2021UnderwaterIE}. Furthermore, it promotes marine tourism, supports scientific research by recording changes in the marine environment, aids in the sustainable management of fisheries, and contributes to educational media \cite{naveen2024advancements}.

The current research in underwater exploration and marine biodiversity plays a critical role in advancing our understanding of the oceans \cite{a6}. Over the past few decades, recent advancements in robotics have led to the widespread use of autonomous underwater vehicles (AUVs) which have enabled deeper and broader exploration of the earth's oceans \cite{jmse10020241}. AUVs rely heavily on visual data to navigate and interpret their surroundings. However, complicated underwater image formation significantly deteriorates underwater images. This has made visual perception for AUVs rather challenging resulting in noisy data collection and relatively challenging applications. One major degradation factor is the changes that occur in the color intensities and visual appearance of the images, which can very significantly according to the type of water, ambient light, object range, and camera response. Such visual anomalies make the correction, enhancement and restoration of underwater images a very challenging process \cite{8578801}.

The formation of underwater images is influenced by the attenuated light that reflects from the scene and the scattering of external light, which reaches the camera sensor and introduces noise. The underwater image formation model is:
\begin{equation}
I = J e^{-\beta^D \cdot z} + B^\infty (1 - e^{-\beta^B \cdot z}),
\end{equation}
where $I$ is the image created at the camera sensor, $J$ is the original unattenuated scene that is being retrieved, $\beta^D$ is the attenuation coefficient, $B^\infty$ is the veiling light, and $\beta^B$ is the backscatter coefficient \cite{8954437}. Estimating the attenuation and backscatter coefficients is a complex task of retrieving the original image $J$ since they are modeled at varying rates across various wavelengths \cite{alma991029871581906387}. 

Similarly, despite the importance of high-quality underwater images, there are several issues and challenges associated with capturing and processing them. This is mainly because underwater scenes include light absorption and scattering effects, which can lead to varying degrees of underwater image quality degradation \cite{a17}. In this regard, we highlight the following key elements that influence the underwater image degradation process \cite{a18}.

\subsection{Sources of Degradation in Underwater Images}
The abundance of suspended particles such as microplankton, dissolved organic matter, etc., in the underwater environment initiates a series of transformations in the propagation direction of light. This means that the light the camera records includes elements of direct reflection, forward scattering, and backscatter due to the random fluctuations in light propagation direction introduced by the suspended particles. More specifically, forward scattering causes light to be deflected from its original path, which results in blurring and a decrease in contrast in the resulting underwater image. Similarly, backscatter produces a marine snow appearance that lowers image quality by reflecting light back toward the camera before it reaches the subject \cite{8578801, 6930829, a19}.

Moreover, the underwater medium significantly distorts images through light attenuation since water absorbs light at different rates depending on wavelength with longer wavelengths facing stronger absorption effects. Similarly, red light fades quickly as depth increases, while blue and green light penetrate farther due to shorter wavelengths. This uneven absorption creates noticeable color distortion in underwater photography making images become darker overall, especially at greater depths. Additionally, the colors shift dramatically as wavelengths disappear at varying rates with the objects far away, starting to lose visibility and contrast. For instance, the camera captures predominantly blue-green scenes instead of the full color spectrum. This loss of light energy results in underexposed images with poor contrast in areas far from the camera \cite{a20, 9711422}.

Another problem is that the artificial light sources reflect off particles between the camera and subject, which creates bright speckles or a milky haze. This interference is particularly problematic in floodlit photography and videography near sediment-rich areas \cite{Li:22, 10.3389/fmars.2023.1163831, isprs-archives-XLVIII-3-2024-1-2024}. The refraction at curves housing ports due to such light sources further magnifies the non-linearity of subjects causing pincushion or barrel distortion. These distortions skew spatial measurements and require calibration for visual applications in marine archaeology or infrastructure inspection \cite{:10.2312/PE/VMV/VMV11/049-056}.

\subsection{Significance of Underwater Image Enhancement for Vision-Based Tasks}
Enhancing underwater images is essential for high-level computer vision tasks such as object detection \cite{a10, a11, a12}, image segmentation \cite{a13, a14}, object tracking \cite{a15}, and classification \cite{a16} in underwater environments. This means that underwater image enhancement (UIE) is a crucial area of study that aims to recover the high-frequency content, which includes fine edges, small textures, and sharp features. The high-frequency details in underwater images refer to the outline of a coral or fish fin, patterns on a shell or ripples in sand, seaweed edges, bubbles, etc. These details are lost due to various degradation factors, such as motion blur, noise, low visibility, and poor contrast \cite{a1}. UIE seeks to improve underwater image quality by mitigating the effects of light scattering, reducing color distortion, enhancing contrast, and restoring fine details. These improvements further facilitate the performance of subsequent high-level vision-related tasks such as image segmentation, object detection, and classification \cite{a2, a3}. Instead of depending on any physical model for the formation of an image, UIE uses qualitative subjective criteria to produce visually appealing underwater images \cite{a4}.

\subsection{Advancements in Underwater Image Enhancement Techniques}
Early UIE methods relied on physical models that focused on accurately calculating medium transmission, and the percentage of scene radiance that reaches the camera \cite{du2024physicalmodelguidedframeworkunderwater} with four main approaches being used, namely polarization, range-gated imaging, fluorescence imaging, and stereo imaging \cite{10.1007/s11036-017-0863-4, TIAN2018515, CHEN20134514, CHEN20142090, Raveendran2021UnderwaterIE}. While these approaches sometimes perform well, they often produced unnatural and poor results in challenging underwater environments \cite{10.1109/TIP.2011.2179666}. Non-physical-based models focus on altering the image's pixel values to improve its visual appeal \cite{Raveendran2021UnderwaterIE}. More recently, deep learning techniques have surpassed traditional physical model-based methods \cite{cong2024comprehensivesurveyunderwaterimage, 9885529}. Simple regression-based convolutional networks have been shown to be effective in basic underwater scenarios. However, they struggle with generalization and robustness in complex scenes. This limitation stems from their inability to fully capture the specific degradation patterns found in real-world underwater conditions.

The latest advancement in the field of image generation, editing, and manipulation involves image-to-image diffusion models, which have shown impressive results in capturing complex image distributions \cite{shi2024cpdmcontentpreservingdiffusionmodel, w16131813, 10812849}. These models enable a smooth transformation between different image domains through iterative refinement steps, offering a promising new approach to tackle these challenging problems \cite{10.1007/s00138-024-01647-8, Croitoru_2023}. They operate on the principle of gradually adding random Gaussian noise to the data during a forward diffusion process and then learning to reverse this process to recover the original data, effectively generating high-quality outputs from pure noise \cite{10.1145/3626235}. Unlike earlier generative adversarial networks (GANs) \cite{IGLESIAS2023100553}, diffusion models provide significantly greater training stability and effectively avoid mode collapse, enabling them to produce more diverse and consistently high-quality outputs \cite{10.1145/3626235}.

The success of image-to-image (I2I) diffusion models in computer vision is exemplified by popular models such as the Diffusion-4K framework \cite{zhang2025diffusion4kultrahighresolutionimagesynthesis}, which enables direct ultra-high-resolution synthesis through wavelet-based fine-tuning. Some other recent I2I models include stable diffusion \cite{zhang2025diffusion4kultrahighresolutionimagesynthesis}, and semantic image synthesis \cite{liu2024iidmimagetoimagediffusionmodel}, which is known for semantic control through cross-attention mechanisms; and I2AM \cite{park2025i2aminterpretingimagetoimagelatent}, which uses bi-directional attribution maps for model debugging and is used for super-resolution and object replacement tasks.

In UIE, the diffusion models face significant limitations due to their dependence on paired datasets. The difficulty in obtaining clear real-world reference images for underwater scenes creates a major obstacle \cite{10220126, bb14, 10196309, tang2023underwaterimageenhancementtransformerbased}. To address this challenge, several researchers have developed methods to synthesize multiple enhanced versions from a single image, then manually select the best-looking result as a reference \cite{8917818, 9930878}. However, this approach introduces subjective preferences into the enhanced results, causing natural domain shift and reduced generalization capability. Similarly, unsupervised or semi-supervised methods offer an alternative that doesn't require paired datasets or relies on minimal paired samples. While these approaches can produce visually appealing underwater images, their performance remains inferior to fully supervised techniques \cite{7995024, a9}.

Another strategy involves leveraging pre-trained diffusion models that contain prior knowledge of transitions between image domains, combined with UIE benchmark datasets to reduce dependency on paired data \cite{DU2025125271}. However, fine-tuning or continued training with these additional datasets risks compromising valuable prior knowledge. Moreover, the absence of real image guidance in the reference domain constrains enhancement quality, often resulting in unrealistic and unnatural enhanced images \cite{10377881}.

Recent research has shown that vision-language models (VLMs) \cite{bordes2024introductionvisionlanguagemodeling, 10445007, ghosh2024exploringfrontiervisionlanguagemodels} contain generic information that can be combined with generative models to create intuitive text-driven interfaces for image generation and manipulation tasks \cite{10.1145/3528223.3530164}, particularly using the contrastive language-image pretraining (CLIP) model  \cite{pmlr-v139-radford21a}. Although CLIP can function adequately as a classifier to differentiate between natural in-air images and degraded underwater photos, directly applying its gradients to control the reverse diffusion process presents significant difficulties. A major challenge lies in identifying precise text prompts that accurately describe various types of underwater degradation. Furthermore, as highlighted in previous research, even minor rephrasing of similar prompts can lead to substantial variations in CLIP scores, creating inconsistency in the enhancement process \cite{Liang2023IterativePL}.

\section{Related Work}

Underwater image enhancement techniques can be broadly classified into two primary categories \cite{jmse10020241}: traditional techniques and deep learning-based techniques. Furthermore, the traditional techniques include physical model-based enhancement methods and non-physical model-based enhancement methods.

\subsection{Physical Models for Underwater Image Enhancement}
The physical model-based enhancement methods uses the physical model of the underwater image formation. They rely on determining the parameters of the physical image formation model, followed by an inverse transformation to convert the underwater image into its corresponding in-air image. 
Some of the well-known physical model-based methods are polarization-based recovery \cite{10.1109/TPAMI.2005.79}, wavelength compensation and dehazing \cite{6104148}, revised underwater image formation model \cite{8578801}, and automatic red-channel underwater image restoration \cite{GALDRAN2015132}. Although these methods are more generalizable than non-physical model-based methods, they require specific operating conditions that match the model being employed. Moreover, these models do not account for human visual perception \cite{10129222}.

\subsection{Non-physical Models for Underwater Image Enhancement}
Non-physical model-based enhancement methods aim to improve the general quality of images without requiring any prior knowledge of the scene environment in certain situations. These methods apply general image processing techniques without explicitly modeling underwater physics. Despite being simpler, image processing-based methods have proven to be ineffective in real-world applications. Variants of histogram equalization, such as contrast limited adaptive histogram equalization (CLAHE) \cite{10.5555/180895.180940}, have gained popularity due to their capacity to improve local contrast. In this context, an integrated color model that independently improves RGB channels to fix color distortion was presented by \cite{5642311}. Similarly, a fusion-based method \cite{6247661} combined multiple versions of an input image to concurrently enhance color balance and contrast. These traditional methods established the groundwork for underwater image enhancement prior to the emergence of learning-based techniques. However, because they disregard the physical image formation model and attenuation factors of the water type, their performance often fails to generalize effectively in complex underwater environments \cite{jmse10020241}.

\subsection{Deep Learning-based Models for Underwater Image Enhancement}
Considering the limitations of traditional enhancement methods, deep learning-based techniques have emerged as a promising alternative. Specifically, in recent years, deep learning has made significant advancements in the field of UIE by outperforming traditional techniques. A plethora of studies have explored different deep learning-based techniques for UIE. Some researchers have deployed convolutional neural networks (CNNs) \cite{bb1, bb2, bb3, bb4, bb6}, while others have developed generative adversarial networks (GANs) \cite{bb7, bb8, pucci2023uwcvganunderwaterimageenhancement, bb10, bb11, 7995024, a9}. Although GAN-based approaches have emerged as a potential solution by generating visually improved underwater images without requiring paired image datasets, these methods tend to be less reliable and generally underperform compared to fully supervised approaches. Furthermore, GAN-based techniques frequently introduce artificial elements and visual artifacts that compromise image authenticity and quality \cite{10.1145/3439723}. They also struggle to capture long-range dependencies effectively. Researchers have also used transformer-based models for UIE \cite{bb12, bb13, bb14, bb15, bb16, bb17, 10484001} due to its ability to capture global context and its notable advancements in various high-level vision tasks. Consequently, deep learning-based techniques are primarily recovery-focused. They lack the flexibility to adapt enhancements based on specific domain needs or contextual variations.

More importantly, deep learning-based techniques rely heavily on large-scale datasets, which are difficult to obtain. Moreover, UIE faces a significant challenge due to the lack of true reference images for comparison. This issue was addressed by developing the underwater image enhancement benchmark (UIEB) \cite{8917818}, where the best outputs from various enhancement approaches were selected to serve as reference standards. However, this approach introduces inherent bias that constrains the effectiveness of current enhancement techniques, particularly deep learning methods, that rely heavily on training data.

In this regard, a promising approach to overcome UIE challenges involves converting in-air natural images into underwater degraded images, creating paired datasets for training. A study \cite{DU2025125271} implemented this concept by applying Koschmieder's light scattering model to transform in-air natural images into pseudo-underwater versions, using the inherent properties of in-air images to mitigate the artificial aspects of synthetic data. However, this physics-based approach often falls short in capturing the full complexity of real-world underwater visual degradation.

Recent research has demonstrated the exceptional capabilities of diffusion models in image generation by reversing a noise process, often modeled using reverse-time stochastic differential equations (SDEs) \cite{gal2023an}. However, this process typically modifies the original image content. To address this limitation, an effective technique has emerged for enhancing the quality of conditional diffusion outputs by using classifier gradients \cite{song2021scorebased, ho2021classifierfree}. Another approach involves establishing connections between input data and conditional information within the model's latent space \cite{9887996, 9879075}. These control mechanisms enable personalization and task-specific image generation, leading to successful application of diffusion models various image enhancement tasks. The image-to-image diffusion framework \cite{9887996} has been adapted for numerous applications, including image enhancement \cite{DU2025125271, 10.1145/3581783.3612378}, inpainting \cite{10.1145/3528233.3530757}, and super-resolution \cite{9887996}.

While diffusion models have shown remarkable progress in generating and enhancing images, they often rely heavily on pixel-level guidance or handcrafted conditions. This highlights the need to explore how effectively vision-language models (VLMs) can be applied to UIE. VLMs have the ability to combine semantic understanding with visual enhancement.

The contrastive language-image pretraining model, otherwise known as CLIP \cite{Radford2021LearningTV}, has demonstrated remarkable capabilities in understanding and processing images in conjunction with textual descriptions. It offers innovative perspectives in image enhancement where both text and image embeddings use the same space, enabling direct comparisons between the two modalities. Its ability to interpret and apply semantic concepts to visual data makes it an ideal candidate for enhancing relevance of data-driven image enhancement methods. The CLIP model consists of a text encoder and an image encoder that are trained together on a vast collection of 400 million curated image-text pairs, it possesses the unique capability to associate visual content with the corresponding textual descriptions. This capability has led to significant advancements in text-to-image generation applications.

Some of the recent applications of CLIP for various tasks are as follows: VQGAN-CLIP \cite{10.1007/978-3-031-19836-6_6} for text-to-image synthesis or artistic image generation, CLIP-LIT \cite{Liang2023IterativePL} for text-to-image generation with literal grounding, and BioViL-T \cite{Bannur2023LearningTE} for medical image analysis. Other studies \cite{10.1145/3528223.3530164, Patashnik_2021_ICCV} have utilized CLIP to modify StyleGAN-generated images using textual prompts. More recently, some studies \cite{9879075, 9879284} have proposed frameworks that combine diffusion models with CLIP guidance to achieve global text-to-image synthesis and enable local image editing. This progression highlights how CLIP is being used to bridge the gap between visual content and contextual understanding, thereby opening new frontiers in image enhancement.

\subsection{Challenges and Limitations}

While text-driven generative models perform exceptionally well with concrete object descriptions such as ``coral,'' ``fish,'' or ``diver,'' they struggle to identify precise prompts for abstract visual qualities and artistic styles. Developing effective prompts for abstract concepts typically requires domain specialists to carefully annotate each image. A significant challenge is the inconsistency in CLIP scores for similar prompts. This limits users' ability to generate and manipulate custom domains for abstract semantic concepts using pre-trained CLIP models, particularly in underwater image enhancement contexts.

To overcome these challenges, recent innovative studies have proposed fine-tuning the CLIP model with substantial training datasets for domain-specific adaptation \cite{10.1145/3528223.3530164}. However, this approach requires significant resources and is time-consuming. An alternative strategy involves prompt learning with a frozen vision encoder to extract precise low-level visual features such as luminance, exposure, and contrast \cite{Liang2023IterativePL, 10.1007/s11263-022-01653-1}. 

Conditional diffusion models excel at mapping transition patterns between diverse image domains \cite{10377881}. Building on this capability, our study trains a conditional diffusion model using a synthetic dataset created through color transfer techniques, enabling it to capture the prior knowledge of mapping transitions between real degraded underwater imagery and real natural in-air scenes. As a more effective alternative to conventional synthesis methods, we employ a color transfer-based strategy \cite{946629} to generate paired training datasets by degrading natural in-air images to closely resemble real underwater visual characteristics. This approach incorporates valuable prior knowledge by establishing direct mapping relationships between real underwater degradation patterns and in-air image properties. We then employ a classifier to preserve this domain knowledge during subsequent fine-tuning of the pre-trained diffusion model, effectively bridging the gap between synthetic training data and real-world underwater images. To achieve this, we use prompt learning techniques to train a classifier capable of distinguishing between natural in-air images and those affected by underwater degradation. The gradients generated by this classifier are used to guide the sampling process of the diffusion model, thereby enhancing its ability to restore underwater imagery more effectively.

\subsection{Contributions}
\begin{figure}[t!]
    \centering
    \includegraphics[width=\columnwidth]{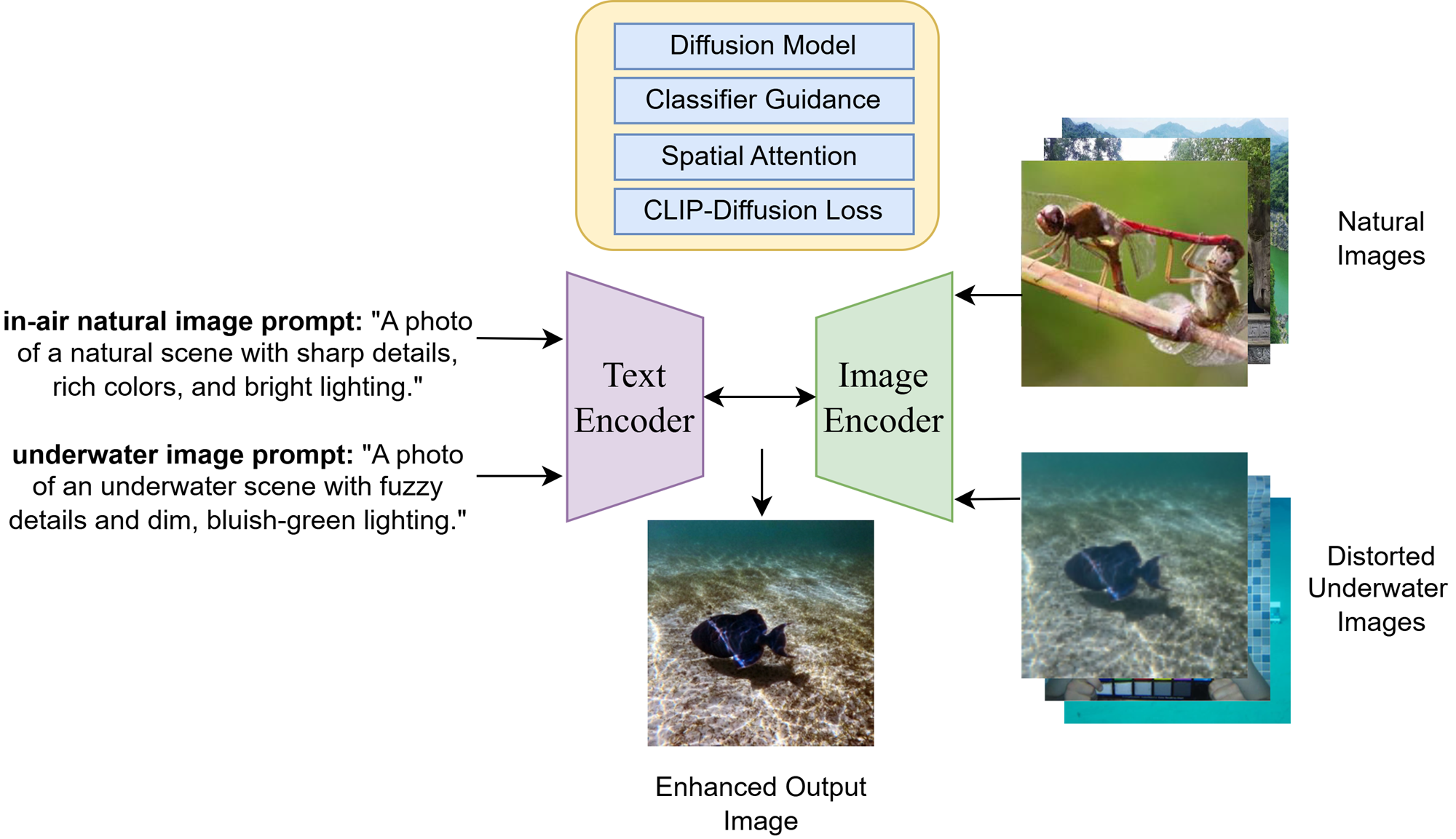}
    \caption{Architecture overview of the proposed UDAN-CLIP model that integrates vision-language alignment and a diffusion model to enhance an underwater image.}
    \label{fig:intro_fig}
\end{figure}

Inspired and motivated by the above-mentioned findings, we use CLIP-UIE \cite{liu2024underwaterimageenhancementdiffusion} as our baseline model and extend it to propose UDAN-CLIP: an Underwater Diffusion Attention Network with Contrastive Language-Image Joint Learning for underwater image enhancement, as illustrated in Figure \ref{fig:intro_fig}. The main features of our model are as follows:
\begin{itemize}
    \item We introduced an underwater diffusion attention network with contrastive language-image joint learning for underwater image enhancement.
    \item We formulated a novel diffusion-aware vision-language loss function that balances pixel-level fidelity, perceptual quality, and semantic alignment to combat color distortion and haze
    \item We integrated a spatial attention mechanism into an image-to-image diffusion model with a contrastive vision-language classifier to strengthen localized feature alignment and boost the performance of underwater image enhancement.
    \item We calculated and compared both full-reference and non-reference image quality metrics to demonstrate that UDAN-CLIP delivers superior performance on real-world underwater image datasets.
\end{itemize}

\section{Proposed Method}
With the scaling victory of customized VLMs paired with the proven effectiveness of diffusion models, we propose UDAN-CLIP: an underwater diffusion attention network. Our approach addresses key challenges, such as, revealing fine details and patterns, correcting color distortion, improving contrast, and enhancing feature resolution in degraded underwater images. We use CLIP-UIE \cite{liu2024underwaterimageenhancementdiffusion}, as our baseline VLM model and further extend it to develop our proposed UDAN-CLIP model.

\subsection{Diffusion Model with Prior Knowledge of Domain Adaptation}
\textit{(a) Color Transfer Synthesis:} A light scattering model can be used to convert natural images to synthetic  underwater images \cite{DU2025125271}. However, these synthetic images have limited variation and degradation types, restricting their ability to represent diverse underwater environments. To accommodate that, color transfer techniques can intuitively and efficiently map color characteristics between image domains \cite{946629}. Therefore, we can transform natural in-air images into near-realistic underwater images with appropriate degradation effects. This means that by providing a wide variety of underwater degradation types, the synthetic dataset becomes more similar to the real underwater domain.
\begin{figure*}[t!]
    \centering
\includegraphics[width=0.9\textwidth]{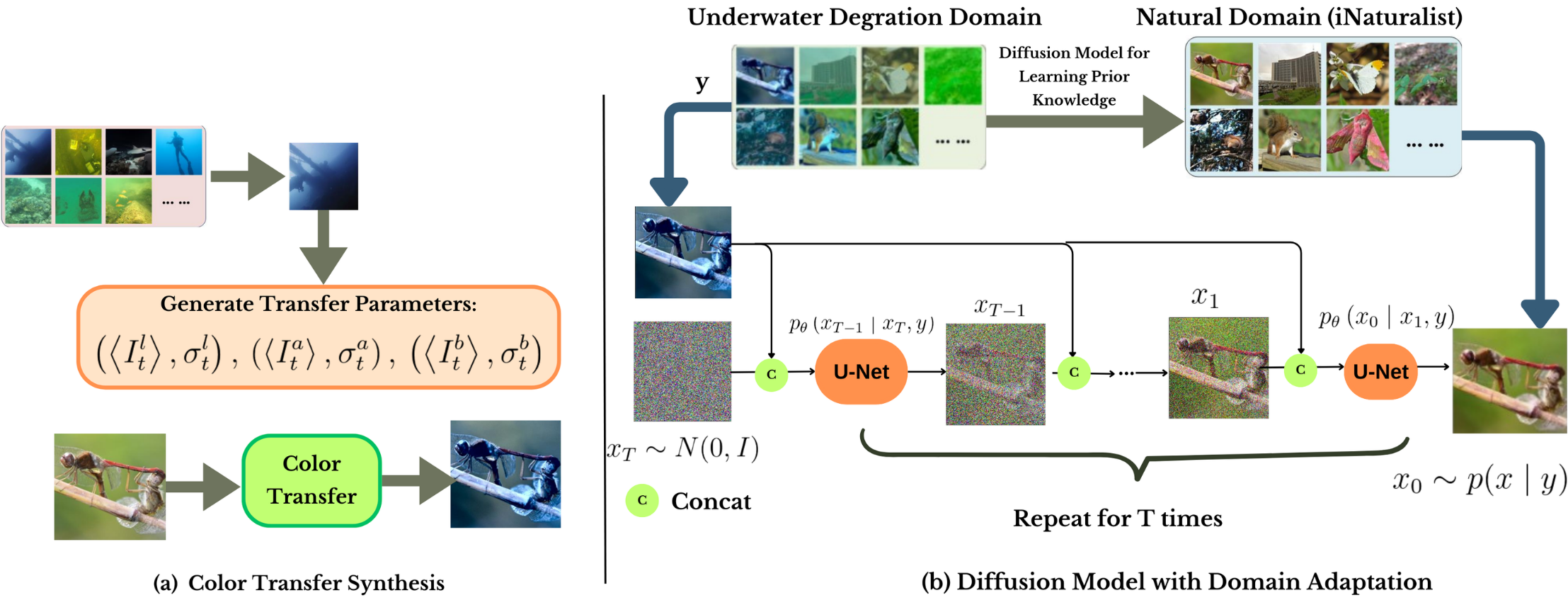}
    \caption{Pre-training Diffusion Model with Prior Knowledge
of Domain Adaption.}
    \label{fig:stage1}
\end{figure*}

To align the color appearance of a source image $I_s$ with a target image $I_t$, we convert all images from RGB to CIELAB color space, $L^*a^*b^*$:

\begin{equation}
\begin{aligned}
l'_s &= \frac{\sigma^{l_t}}{\sigma^{l_s}} (l_s - \langle I^{l_s} \rangle) + \langle I^{l_t} \rangle, \\
a'_s &= \frac{\sigma^{a_t}}{\sigma^{a_s}} (a_s - \langle I^{a_s} \rangle) + \langle I^{a_t} \rangle, \\
b'_s &= \frac{\sigma^{b_t}}{\sigma^{b_s}} (b_s - \langle I^{b_s} \rangle) + \langle I^{b_t} \rangle,
\end{aligned}
\label{eq:color_transfer}
\end{equation} where $\langle \cdot \rangle$ refers to the calculation for the mean value of the channel, and $\sigma_s^i$, $\sigma_t^i$, $i \in [l, a, b]$ are the standard deviations of the different channels in the source and target images, respectively. The terms $l_s$, $a_s$, and $b_s$ are the pixel values of the source image, while $l'_s$, $a'_s$, and $b'_s$ are the pixel values of the resulting image. 

In-air natural images are taken in natural air such as on land or above the water surface. These images are clear, undistorted, and well-lit images captured. In our proposed approach, an underwater image is randomly selected from the template pool for each in-air natural image $I_s$. where the template pool is considered as  the underwater scene domain. The randomly selected image serves as both the target image $I_t$ and the template. This $I_t$ guides the transition of $I_s$ into the underwater domain through Eq. \eqref{eq:color_transfer} as illustrated in detail in Figure \ref{fig:stage1}. After this transformation, the synthetically degraded image is reconverted to original RGB color space. We refer to the color transfer-generated synthetic dataset is known as as UIE-air, whereas the manually curated synthetic dataset is termed as UIE-ref, which includes benchmark collections such as UIEB \cite{8917818} and SUIM-E \cite{9930878}.

\textit{(b) Diffusion Model with Domain Adaptation:}
With the paired UIE-air dataset $D = \{(x_i, y_i)\}_{i=1}^N$, the objective is to to train a conditional diffusion model that encapsulates the transitional mapping knowledge from underwater degradation to real in-air natural underwater domain. The dataset $\mathcal{D} = \{(x_i, y_i)\}_{i=1}^N$ comprises paired samples where,  $x_i$ represents the synthetic underwater degraded image generated via color transfer from an in-air natural image, while $y_i$ denotes the original in-air natural image that is the ground truth before degradation. For this case, we reformulate our optimization objective as:
\begin{equation}
\label{eq:3}
\mathcal{L}_2 = \min_{\theta} \mathbb{E}_{x_0 \sim q(x_0), \, \epsilon \sim \mathcal{N}(0, \mathbf{I}), \, y, \, t} 
\left\| \epsilon - \epsilon_{\theta}(x_t, y, t) \right\|^2, 
\end{equation}
This is accomplished by minimizing the learning objective $\mathcal{L}_2$ as per Eq. \ref{eq:3} for an in-air natural image $x$, conditioned on a corresponding synthesized underwater image $y$. Minimizing $\mathcal{L}_2$ trains the diffusion model to predict noise $\epsilon$ during the forward process, conditioned on the degraded input $y$, noisy state $x_t$, and timestep $t$. This preserves prior domain transitions (underwater$\rightarrow$in-air) and uses multi-guidance to align results with structural fidelity and natural appearance. Focusing on high-frequency regions accelerates fine-tuning, preventing catastrophic forgetting and ensuring realistic enhanced images. As shown in Figure \ref{fig:stage1}, the classifier $p(y|x_t)$ provides gradients that directly steer the reverse diffusion process, ensuring sample $x_t$ maintains consistency with the conditioning information $y$. Once training is completed, this means that the learned prior knowledge is used to counteract negative effects from synthetic data artifacts.

\subsection{Classifier-Guided Knowledge Preservation}
We use our pre-trained model in a subsequent step to fine-tune it on the UIE-ref dataset for addressing specific underwater degradation scenarios. However, conventional fine-tuning or continued training approaches risk deteriorating the previously acquired knowledge, potentially resulting in catastrophic forgetting and the collapse of the model. To overcome this challenge, we propose a classifier-guided framework by training a domain classifier that encapsulates prior knowledge of in-air natural images, effectively discriminating between in-air natural images and underwater domain images. Subsequently, this classifier is employed along with the UIE-ref dataset during the diffusion model fine-tuning process. This strategic implementation preserves the essential prior knowledge throughout fine-tuning for adaptation to specific underwater conditions.
The score function of the multicondition
model can be derived from the Bayes formula as:
\begin{equation}
\label{4}
\begin{split}
\nabla \log p(x_t|y_1, \ldots, y_m) &= \nabla \log \left( \frac{p(x_t)p(y_1, \ldots, y_m)}{p(y_1, \ldots, y_m|x_t)} \right) \\
&= \nabla \log p(x_t) + \nabla \log p(y_1, \ldots, y_m|x_t),
\end{split}
\end{equation}
where the conditions are independent of each other is an assumption. Furthermore, the Eq. \ref{5} is derived as follows:
\begin{equation}
\label{5}
\begin{split}
\nabla \log p(y_1, y_2 | x_t) 
&= \nabla \log p(y_1|x_t) + \cdots + \nabla \log p(y_m|x_t).
\end{split}
\end{equation}
Since there are only two conditions, the source image $y_1$ and the in-air natural domain $y_2$ in this UIE task, the Eq. \ref{5} has been simplified by setting $m=2$. Then, substituting this into Eq. \ref{4}, gives us:
\begin{equation}
\label{6}
\begin{split}
\nabla \log p(x_t|y_1, y_2) 
&= \nabla \log p(x_t) + \nabla \log p(y_1|x_t) \\
&\quad + \nabla \log p(y_2|x_t),
\end{split}
\end{equation}
where $p(y_1|x_t)$ is the same classifier as $p(y|x_t)$ as expressed below: 
\begin{equation}
\label{7}
\begin{aligned}
\nabla \log p(x_t \mid y) &= \nabla \log \left( \frac{p(x_t) p(y \mid x_t)}{p(y)} \right) \\
&= \nabla \log p(x_t) + \nabla \log p(y \mid x_t) - \nabla \log p(y) \\
&= \nabla \log p(x_t) + \nabla \log p(y \mid x_t).
\end{aligned}
\end{equation}
This makes the sample $x_t$ adhere to the source image $y_1$. The $p(y_2|x_t)$ is the trained classifier that makes the sample $x_t$ move towards the in-air natural domain $y_2$.

To enable fine-grained control over the model's tendency toward or away from the in-air natural domain, we introduce a hyperparameter term $\lambda \in [0, 1]$. This is done to ensure that adversarial gradient of the classifiers can be scaled. This is shown in the Eq. \ref{8} below:
\begin{align}
\nabla \log p(x_t \mid y_1, y_2) &= \nabla \log p(x_t) 
+ \lambda \nabla \log p(y_1 \mid x_t) \nonumber \\
&\quad + (1 - \lambda) \nabla \log p(y_2 \mid x_t).
\label{8}
\end{align}
When $\lambda < 0.5$, the conditional diffusion model gives preference to the conditioning information $y_2$. This results in generated data that aligns more closely with the in-air natural domain illustrated in Figure \ref{fig:stage1}. In accordance with Eq. \ref{eq:3} and Eq. \ref{9}, the noise prediction function  $\epsilon_\theta(x_t, t)$ is rewritten to incorporate both classifier gradient. As per \cite{song2020score, 10.5555/3454287.3455354} diffusion models are unified by reformulating the discrete denoising process as a continuous stochastic differential equation (SDE), where the reverse process simulates score functions $\nabla \log p(x_t)$ from time $T$ to $0$. They demonstrate that score matching is equivalent to ELBO optimization and establish a linear relationship between the score function $\nabla \log p(x_t)$ and the noise prediction network $\epsilon_\theta(x_t, t)$, creating a mathematical bridge between discrete diffusion models and continuous score-based generative approaches. This relationship is shown in the Eq. \ref{9} below:
\begin{equation}
\label{9}
\nabla \log p(x_t) = -\frac{1}{\sqrt{1 - \bar{\alpha}_t}} \, \epsilon_\theta(x_t, t).
\end{equation}
The noise prediction function  $\epsilon_\theta(x_t, t)$ is rewritten to incorporate both classifier gradients as $\epsilon_\theta(x_t, y_1, y_2, t)$ as:
\begin{equation}
\label{10}
\begin{split}
\epsilon_\theta(x_t, y_1, y_2, t) &= \epsilon_\theta(x_t, t) 
- \alpha \sqrt{1 - \bar{\alpha}_t} \nabla \log p(y_1 | x_t) \\
&\quad - (1 - \alpha) \sqrt{1 - \bar{\alpha}_t} \nabla \log p(y_2 | x_t).
\end{split}
\end{equation}
The learning objective of this complete multi-guidance diffusion model is $\mathcal{L}_{\text{multi}}$ as expressed below:  
\begin{equation}
\label{11}
\mathcal{L}_{\text{multi}} = \min_{\theta} \mathbb{E}_{x_0 \sim q(x_0), \epsilon \sim \mathcal{N}(0, \mathbf{I}), y_1, y_2, t} \left\| \epsilon - \epsilon_\theta(x_t, y_1, y_2, t) \right\|^2.
\end{equation}
The learning objective $\mathcal{L}_{\text{multi}}$ is applied directly during the fine-tuning process of our proposed diffusion model. Through multi-condition guidance, the denoising trajectory from $x_t$ to $x_0$ remains constrained by both the source image $y_1$ and the in-air natural domain $y_2$. This minimizes the potential deterioration that fine-tuning can cause to the prior knowledge
of the pre-trained model.

\subsection{Prompt Learning for Domain-Adaptive Classification}
In our proposed approach, we construct two specialized classifiers: $p(y_2|x_t)$ with conditional information $y_2$ representing the in-air natural domain, and $p(y_1|x_t)$ with conditional information $y_1$ corresponding to the UIE-ref dataset. When combined, these classifiers jointly align the diffusion model's generation process, effectively steering results toward the in-air natural domain.

To implement this, we propose a VLM-based classifier. Specifically, we use CLIP-UIE as our baseline to utilize its promising semantic understanding \cite{pmlr-v139-radford21a}. The central task of our proposed classifier is to identify prompts that can effectively distinguish between in-air natural images and underwater images. However, one significant challenge is that the prompts for quality and abstract perception of these image types demand extensive manual prompt tuning, with similar prompts often producing dramatically different classification scores.

To address this limitation, the CoOp \cite{10.1007/s11263-022-01653-1} methodology offers a solution by representing the context words of prompts as learnable vectors, thereby enabling VLMs to be adapted effectively for the downstream tasks. In our model, only two prompts are required to characterize in-air natural and underwater images. This motivation comes from CLIP-LIT's \cite{Liang2023IterativePL} approach, where prompt learning has been utilized to model a prompt text with the longest learnable tensors, ultimately maximizing image characterization capabilities.
It is important to note that throughout this entire process, our baseline model remains frozen. We further divide our prompt learning process with two components: prompt initialization and prompt training as explained in the subsequent sections.

{\textit{(a) Prompt Initialization:}}
The initialization process begins with arbitrarily selected textual descriptions for both in-air natural and underwater images. Our baseline model transforms these descriptions into the multi-modal embedding space, generating two distinct prompt representations: the in-air natural image prompt $T_n \in \mathbb{R}^{N \times M}$ and the underwater image prompt $T_u \in \mathbb{R}^{N \times M}$. Here, $N$ denotes the token length of each prompt, which has a maximum limit of 77 tokens. Based on the existing initializations from CoOp \cite{10.1007/s11263-022-01653-1}, performance improvements are positively correlated with an increase in the number of prompt tokens. Consequently, $T_n$ and $T_u$ as learnable tensors have been designed with $N = 77$, thereby maximizing the available prompt token length for optimal representational capacity.

{\textit{(b) Prompt Training:}}
Given an in-air natural image \( I_n \) and an underwater image \( I_u \), the spatial attention-guided encoder processes these images to extract hierarchical features. These features are refined via attention masking (as mentioned in the next subsection) and pooled into a global descriptor \( \phi(I) \), which is then aligned our model's text embeddings. Concurrently, the prompts $T_n$ and $T_u$ are fed into the text encoder $\Theta(\cdot)$ to derive their textual embeddings. For a batch containing image-text pairs, the prediction probability is calculated as:
\begin{equation}
\label{12}
P(T_t \mid I) = \frac{e^{\cos(\Theta(T_t), \phi(I))}}{\sum_{i \in \{n, u\}} e^{\cos(\Theta(T_i), \phi(I))}},
\end{equation}
where \( \phi(I) \) denotes the attention-refined features, $I$ represents an image from the set $\{I_n, I_u\}$, and $\cos(\cdot)$ denotes the cosine similarity. Notably, the textual prompts $T_n/T_u$ are shared among all in-air natural/underwater images. The prompt learning methodology used in this study aims to align the learnable prompts with their corresponding images in the embedding space. This is achieved by maximizing the cosine similarity for matching pairs $\{T_n, I_n\}$ and $\{T_u, I_u\}$, while simultaneously minimizing similarity for non-matching pairs. Since the prompt learning in our model involves only two prompts, the process can be streamlined by directly implementing binary cross-entropy for classifying in-air natural and underwater images, thereby facilitating the learning of prompt tensors as expressed:

\begin{equation}
\label{13}
\mathcal{L}p = -(q \ast \log P(T_n \mid I) + (1 - q) \ast \log(1 - P(T_n \mid I))), 
\end{equation}
where $q$ represents a one-hot label that equals 1 for in-air natural images and 0 for underwater images.

\subsection{Spatial Attention in UDAN-CLIP}

\begin{figure*}[t!]
    \centering
\includegraphics[width=0.9\textwidth]{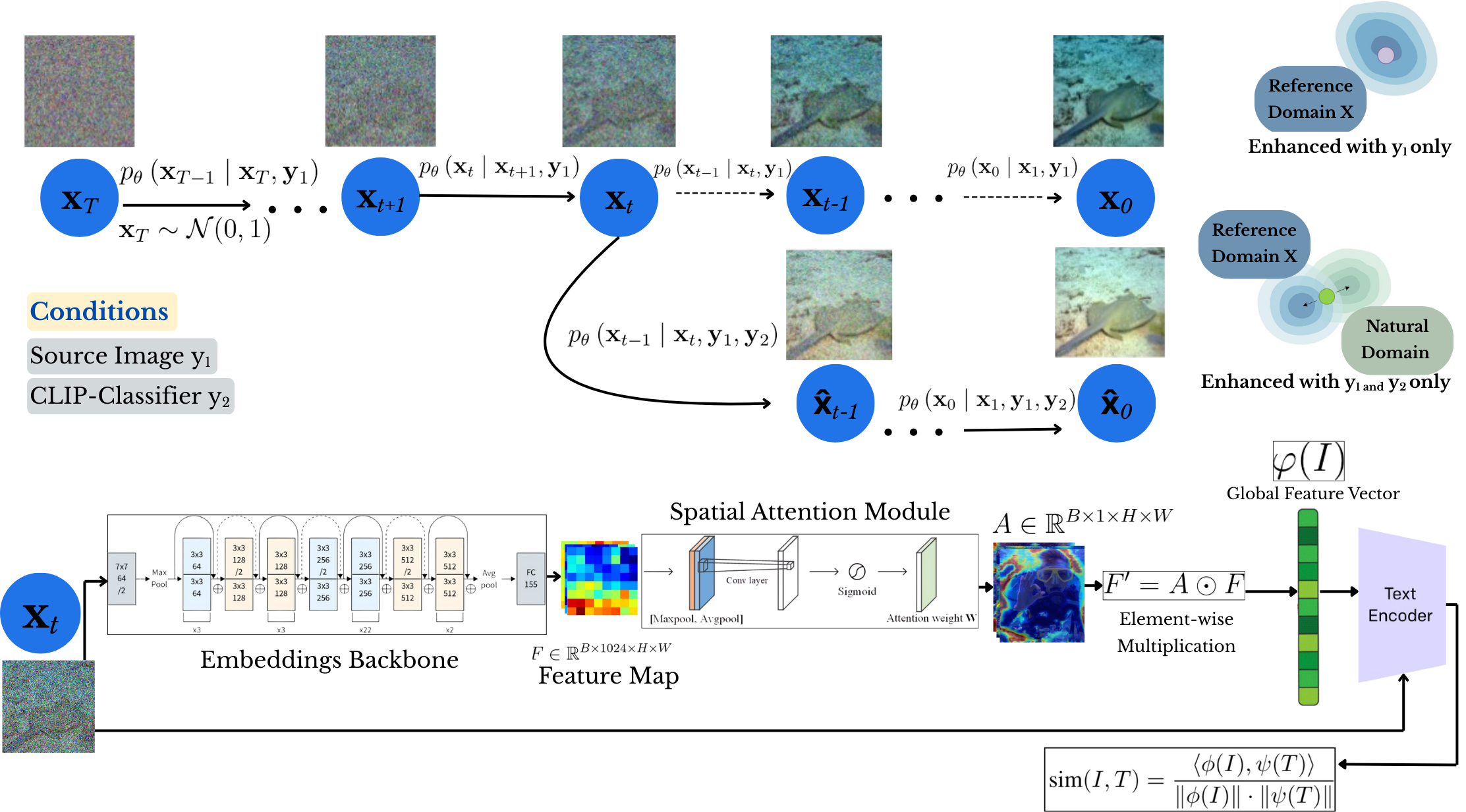}
    \caption{Fine-tuning UDAN-CLIP with spatial attention and multi-classifier guidance strategy on real-underwater datasets. The intermediate results from $x_t \ \text{to} \ x_0$ move towards the in-air natural domain, mitigating the damage of fine-tuning to the prior knowledge of the pre-trained model.}
    \label{fig:stage2}
\end{figure*}

In our proposed UDAN-CLIP model, we proposed  spatial attention as shown in Figure \ref{fig:stage2} to address localized degradations such as haze, low contrast, turbidity. Turbidity refers to the cloudiness or haziness in water caused by suspended particles like sediments, plankton, or organic matter. The backbone encoder of our model yields a feature map \( F \in \mathbb{R}^{B \times 1024 \times H \times W} \), preserving fine-grained spatial hierarchies critical for local degradation detection. Additionally, a learnable attention module processes \( F \) to generate a spatially weighted 1-channel attention mask \( A \in \mathbb{R}^{B \times 1 \times H \times W} \). This mask is an element-wise multiplied with the original features to suppress irrelevant regions and amplify degraded areas.
The refined features \( F' \) are computed as:

\begin{equation}
\label{16}
F' = A \odot F,
\end{equation} where \( \odot \) denotes element-wise multiplication. The attended features \( F' \) are pooled into a global descriptor using adaptive attention pooling, by combining adaptive average pooling with a learned projection. 
This produces a normalized feature vector \( \phi(I) \), which is aligned with the text embeddings via cosine similarity as expressed below:

\begin{equation}
\label{17}
\text{sim}(I, T) = \frac{\langle \phi(I), \psi(T) \rangle}{\|\phi(I)\| \cdot \|\psi(T)\|}.
\end{equation}

With this strategy, attention maps reveal focused activation on turbid zones, enabling context-aware restoration. The attention-refined features directly influence the loss component of our proposed novel loss function as explained in the subsequent sections. By aligning attended regions with semantic prompts (for instance, ``vibrant coral reef''), UDAN-CLIP ensures that diffusion steps are semantically guided to reconstruct critical areas such as sharpening edges of marine life. This also ensures that UDAN-CLIP bridges pixel-level restoration and semantic guidance, which is particularly effective for localized underwater degradations.

\subsection{Joint Visual-Textual Alignment}
To facilitate joint visual-textual alignment, we obtain the optimized prompts with out VLM-based classifier. This classifier operates by measuring the alignment between visual features of the generated image and the textual embeddings of the trained prompts within the embedding space of our model as expressed below:
\begin{equation}
\mathcal{L}_{\text{classifier}} = \frac{e^{\cos(\Psi(T_u), \phi(I_g))}}{\sum_{i \in \{n,u\}} e^{\cos(\Psi(T_i), \phi(I_g))}},
\label{14}
\end{equation}
where $I_g$ represents the image produced by the diffusion model, and \( \phi(I_g) \) denotes the attention-refined features.
By incorporating this $\mathcal{L}_{\text{classifier}}$ into Eq. \ref{10}, the noise prediction function $\epsilon\theta(x_t, y_1, y_2, t)$ can be reformulated and expressed below:
\begin{equation}
\begin{split}
\epsilon_{\theta}(x_t, y_1, y_2, t) =\ & \epsilon_{\theta}(x_t, t) 
- \gamma_1 \sqrt{1 - \bar{\alpha}_t} \nabla \log p(y_1 \mid x_t) \\
& - \gamma_2 \sqrt{1 - \bar{\alpha}_t} \nabla \mathcal{L}_{\text{CLIP-Diff}}
\end{split},
\label{15}
\end{equation}where $y_2$ = $T_u$ due to the inherent properties of the cosine function.
    
\subsection{CLIP-Diffusion Loss}
To further enhance visual quality and semantic consistency, we propose a novel loss function $\mathcal{L}_{\text{UDAN-CLIP}}$. The $\mathcal{L}_{\text{UDAN-CLIP}}$ loss is used to fine-tune UDAN-CLIP on real-world underwater datasets by integrating the standard diffusion noise prediction loss with a perceptual alignment term derived from the embedding of our proposed UDAN-CLIP model. By using the rich, pre-trained semantic representations of our model, the perceptual component encourages the generated images to match the target distribution in pixel space and align with high-level semantic features. This is especially critical in the UIE task, where conventional losses may fail to capture perceptual fidelity due to color shifts and structural degradation. The proposed $\mathcal{L}_{\text{UDAN-CLIP}}$ effectively balances low-level reconstruction and high-level semantic guidance, leading to enhanced detail preservation, improved color correction, and better alignment with human perception. Our proposed $\mathcal{L}_{\text{UDAN-CLIP}}$ is formulated as follows:   

\begin{equation}
\label{200}
\begin{split}
\mathcal{L}_{\text{UDAN-CLIP}} =\ & 
\lambda_1 \cdot \mathbb{E}_{\mathbf{x}, \boldsymbol{\epsilon}} \left[ 
\left\lVert \boldsymbol{\epsilon} - \boldsymbol{\epsilon}_\theta(\mathbf{x}) \right\rVert_1 
\right] \\
& + \lambda_2 \cdot \mathbb{E}_{\mathbf{x}} \left[ 
\mathcal{D}_{\text{CLIP}} \left( f_\theta(\mathbf{x}), f_\theta(\mathbf{x}_{\text{target}}) \right) 
\right]
\end{split},
\end{equation}
\noindent
where $\boldsymbol{\epsilon}$ is the ground truth noise, $\boldsymbol{\epsilon}_\theta(\mathbf{x})$ is the model's predicted noise for input $\mathbf{x}$, $f_\theta(\cdot)$ is the UDAN-CLIP encoder, and $\mathcal{D}_{\text{CLIP}}(\cdot, \cdot)$ denotes a perceptual distance in the embedding space. We use the two scalar weights, $\lambda_1$ and $\lambda_2$, to balance the contributions of the diffusion loss and the perceptual alignment loss of our UDAN-CLIP model, respectively. The value of $\lambda_1$ controls the influence of the diffusion process in guiding visual generation or alignment, while $\lambda_2$ adjusts the emphasis on aligning visual features with textual representations through UDAN-CLIP. The choice of these weights directly affects the trade-off between generative fidelity and semantic consistency.

\section{Experiments}
\subsection{Datasets}

The acquisition of real reference images for UIE is a well-known challenge in this field. As part of stage-one of our methodology, color transfer techniques were applied to transform in-air natural images from the INaturalist 2021 dataset \cite{8579012} into synthetic underwater versions. This creates paired samples known as the UIE-air dataset as shown in Figure \ref{fig:stage1}.

Furthermore, we mainly use two widely used real-world UIE datasets. The SUIM-E \cite{9930878} dataset contains 1, 635 images, while the UIEB \cite{8917818} consists of 890 real-world underwater images. The UIEB dataset covers various scene categories such as coral reefs and marine life, and is further divided into two subsets. The first subset contains 890 raw images with corresponding high-quality reference images. The second subset has 60 challenging images without satisfactory reference images. These images suffer from issues such as color casts, low contrast, and blurring. Our proposed UDAN-CLIP model is fine-tuned using the SUIM-E-train set, which has 1, 525 paired underwater images, in addition to 800 paired images that were chosen at random from UIEB, as the training set. A test set of 200 images, known as T200, was created by combining the test set from SUIM-E and the remaining images of UIEB.

Moreover, we also evaluate the robustness and generalization of our model on additionally widely used datasets of underwater images that lack reference images images like the Color-Checker7 \cite{gsciede00note:crna03} dataset which includes seven underwater images captured in a shallow swimming pool using various cameras. This dataset is used to assess the performance of the proposed UDAN-CLIP for color correction. More experiments were performed on the C60 \cite{8917818} dataset, which consists of 60 challenging enhancement cases without reference images. These 60 images are considered particularly difficult because they exhibit severe degradation, such as strong color casts, extremely low brightness,  and poor visibility. These conditions make it very difficult to obtain satisfactory reference images for this dataset.

The INaturalist 2021 dataset encompasses 500,000 images distributed across 10,000 categories, with 50 images per category. The image-to-image diffusion model of UDAN-CLIP was trained using this dataset to learn prior knowledge of mapping transitions from the real underwater degradation domain to the in-air natural domain.

\subsection{Implementation Details}

We implement our UDAN-CLIP model using the PyTorch framework. For training, we employ the Adam optimizer with an initial learning rate of $1 \times 10^{-6}$, accompanied by a scheduler that linearly decays the learning rate to zero. The $\mathcal{L}_{\text{UDAN-CLIP}}$ loss weights $\lambda_1 = 0.6$ and $\lambda_2 = 0.4$ are empirically set to balance denoising and perceptual alignment. The UDAN-CLIP encoder $f_{\theta}(\cdot)$ leverages ResNet-101 with spatial attention for feature extraction. The diffusion process is carried out over 2000 time steps ($T = 2000$) using a linear noise schedule ranging from $1 \times 10^{-6}$ to $1 \times 10^{-2}$. Furthermore, the training data undergoes standard data augmentation techniques including random rotations and horizontal flips. The experiments were conducted on a consistent hardware setup comprising of NVIDIA GeForce RTX 3090 GPU. The pre-trained diffusion model that was utilized in the UDAN-CLIP model was trained from scratch for 300 hours on the UIE-air dataset \cite{liu2024underwaterimageenhancementdiffusion}.

\subsection{Comparison with the State-of-the Art Methods}

We compare our proposed UDAN-CLIP model with the state-of-the-art methods such as UDCP \cite{6755982} and ULAP \cite{10.1007/978-3-030-00776-8_62}, Ucolor \cite{9426457}, TCTL-Net \cite{10298280}, UIEC\textsuperscript{2}-Net \cite{WANG2021116250}, UDAformer \cite{SHEN202377}, DM\_underwater \cite{10.1145/3581783.3612378}, CLIP-UIE \cite{liu2024underwaterimageenhancementdiffusion}). The first two methods UDCP \cite{6755982} and ULAP \cite{10.1007/978-3-030-00776-8_62} that have been used for quantitative comparisons are the most representative traditional methods that have gained widespread use in recent years. The six deep learning-based methods under comparison have attained state-of-the-art (SOTA) performance in the past years. Among these, DM\_underwater \cite{10.1145/3581783.3612378} is an early UIE technique based on a diffusion model, while UDAformer \cite{SHEN202377}, uses a dual attention transformer structure to create the enhancing framework. More importantly, the same testing set was used from the SUIM-E and UIEB datasets to test all deep learning-based methods in order to guarantee a rigorous and fair comparison. Furthermore, we use the same experimental protocols for all the comparisons.
 
\subsection{Quantitative Evaluations}
For quantitative evaluation, we use two reference metrics: peak signal to noise ratio (PSNR) \cite{a65} and structural similarity index measure (SSIM) \cite{a66}. They are used to assess and measure the consistency between the enhanced images and ground truth images. We also use four non-reference metrics including underwater image quality measure (UIQM) \cite{a66}, underwater image sharpness measure (UISM), and naturalness image quality evaluator (NIQE). For all these metrics, higher values indicate better performance. We show a summary of the quantitative results in Table \ref{tab:Quantitative_comp}, covering the three most widely used real-world datasets: T200, Color-Checker7, and C60. Regarding the results on the T200 dataset, as presented in Table \ref{tab:Quantitative_comp}, our UDAN-CLIP model outperforms the baseline method CLIP-UIE \cite{liu2024underwaterimageenhancementdiffusion} in key metrics such as PSNR, SSIM, and UCIQE, demonstrating superior performance in image quality and structural preservation. While CLIP-UIE shows slightly better results in UIQM and CPBD, the differences are minimal. Overall, the balanced performance across all metrics suggests that UDAN-CLIP effectively enhances image quality while maintaining structural integrity and improving perceptual sharpness and color fidelity.

Our proposed UDAN-CLIP model achieves the highest PSNR score of 27.949, demonstrating superior overall image enhancement quality. This reflects the model’s ability to generate reconstructions that are more faithful to the ground truth. While DM\_underwater and UDAformer perform competitively, they do not reach the level of enhancement fidelity achieved by UDAN-CLIP. Traditional methods such as UDCP and ULAP yield significantly lower PSNR values, highlighting the clear advantage of our deep learning-based framework. In terms of structural similarity, UDAN-CLIP again outperforms all competing methods with a leading SSIM score of 0.952, indicating excellent preservation of fine details and image structure. DM\_underwater and UDAformer achieve values of 0.931 and 0.921, respectively, and the baseline model CLIP-UIE, while strong at 0.936, still lags behind our method. For UIQM, which assesses perceptual quality in underwater conditions, CLIP-UIE reports the highest score of 0.981, with UDAN-CLIP closely behind at 0.961. Although this margin is narrow, UDAN-CLIP demonstrates more consistent and balanced performance across all other evaluation metrics. Traditional methods such as UDCP, despite achieving a score of 1.268, show limited overall enhancement quality compared to deep learning-based approaches. UDAN-CLIP also achieves the highest UCIQE score of 0.654, indicating superior enhancement of color and contrast in underwater scenes. Competing methods show comparatively lower performance in this metric, while traditional techniques fall significantly short, underscoring UDAN-CLIP’s strength in visual enhancement. Although UDCP records the highest CPBD value of 0.636, which suggests greater sharpness, this often compromises other important visual attributes. UDAN-CLIP achieves a competitive CPBD score of 0.615, offering a more balanced outcome by maintaining perceptual sharpness alongside superior color fidelity, structural integrity, and overall enhancement quality.
\begin{landscape}
\newcommand{\gup}{\textcolor{customgreen}{$\uparrow$}} 
\newcommand{\rdown}{\textcolor{red}{$\downarrow$}}     
\begin{table*}[t!]
\centering
\begin{minipage}{\linewidth}
\caption{A quantitative comparison of underwater image enhancement methods is conducted on the T200, Color-Checker7, and C60 datasets. The reported values in brackets indicate the difference from our proposed model UDAN-CLIP. A green upward arrow (↑) denotes an improvement achieved by UDAN-CLIP over the competing methods, while a red downward arrow (↓) indicates a slight performance drop compared to these methods. Positive differences (shown in green with ↑) reflect cases where UDAN-CLIP outperforms others, whereas negative differences (shown in red with ↓) highlight instances where UDAN-CLIP falls slightly behind. The best-performing result for each metric across all methods is highlighted in blue to indicate the highest overall performance.}
\label{tab:Quantitative_comp}
\resizebox{\textwidth}{!}{
\begin{tabular}{lccccccccccc}
\toprule
Methods & \multicolumn{5}{c}{T200 \cite{liu2024underwaterimageenhancementdiffusion}} & \multicolumn{3}{c}{Color-Checker7 \cite{gsciede00note:crna03}} & \multicolumn{3}{c}{C60 \cite{8917818}} \\
\cmidrule(lr){2-6} \cmidrule(lr){7-9} \cmidrule(lr){10-12}
 & PSNR~(\color{customgreen}$\uparrow\color{red}\downarrow$\color{black}) & SSIM~(\color{customgreen}$\uparrow\color{red}\downarrow$\color{black}) & UIQM~(\color{customgreen}$\uparrow\color{red}\downarrow$\color{black}) & UCIQE~(\color{customgreen}$\uparrow\color{red}\downarrow$\color{black}) 
 & CPBD~(\color{customgreen}$\uparrow\color{red}\downarrow$\color{black}) & UIQM~(\color{customgreen}$\uparrow\color{red}\downarrow$\color{black}) & UCIQE~(\color{customgreen}$\uparrow\color{red}\downarrow$\color{black}) 
 & CPBD~(\color{customgreen}$\uparrow\color{red}\downarrow$\color{black}) & UIQM~(\color{customgreen}$\uparrow\color{red}\downarrow$\color{black}) & UCIQE~(\color{customgreen}$\uparrow\color{red}\downarrow$\color{black}) & CPBD~(\color{customgreen}$\uparrow\color{red}\downarrow$\color{black}) \\
\midrule
UDCP \cite{6755982} & 11.803 (\textcolor{customgreen}{\gup 16.146}) & 0.548 (\textcolor{customgreen}{\gup 0.404}) & \textcolor{blue}{1.268} (\textcolor{red}{\rdown 0.307}) & 0.598 (\textcolor{customgreen}{\gup 0.056}) & \textcolor{blue}{0.636} (\textcolor{red}{\rdown 0.021}) & \textcolor{blue}{1.642} (\textcolor{red}{\rdown 0.337}) & 0.641 (\textcolor{customgreen}{\gup 0.042}) & 0.589 (\textcolor{red}{\rdown 0.045}) & \textcolor{blue}{0.952} (\textcolor{red}{\rdown 0.268}) & 0.552 (\textcolor{customgreen}{\gup 0.057}) & \textcolor{blue}{0.551} (\textcolor{red}{\rdown 0.043}) \\
ULAP \cite{10.1007/978-3-030-00776-8_62} & 16.570 (\textcolor{customgreen}{\gup 11.379}) & 0.768 (\textcolor{customgreen}{\gup 0.184}) & 0.955 (\textcolor{customgreen}{\gup 0.006}) & 0.614 (\textcolor{customgreen}{\gup 0.040}) & 0.647 (\textcolor{red}{\rdown 0.032}) & 0.752 (\textcolor{customgreen}{\gup 0.553}) & 0.628 (\textcolor{customgreen}{\gup 0.055}) & 0.557 (\textcolor{red}{\rdown 0.013}) & 0.690 (\textcolor{red}{\rdown 0.006}) & 0.579 (\textcolor{customgreen}{\gup 0.030}) & 0.560 (\textcolor{red}{\rdown 0.052}) \\
Ucolor \cite{9426457} & 21.907 (\textcolor{customgreen}{\gup 6.042}) & 0.888 (\textcolor{customgreen}{\gup 0.064}) & 0.586 (\textcolor{customgreen}{\gup 0.375}) & 0.587 (\textcolor{customgreen}{\gup 0.067}) & 0.606 (\textcolor{customgreen}{\gup 0.009}) & 0.693 (\textcolor{customgreen}{\gup 0.612}) & 0.586 (\textcolor{customgreen}{\gup 0.097}) & 0.580 (\textcolor{red}{\rdown 0.036}) & 0.505 (\textcolor{customgreen}{\gup 0.179}) & 0.553 (\textcolor{customgreen}{\gup 0.056}) & 0.512 (\textcolor{red}{\rdown 0.004}) \\
TCTL-Net \cite{10298280} & 22.403 (\textcolor{customgreen}{\gup 5.546}) & 0.897 (\textcolor{customgreen}{\gup 0.055}) & 0.796 (\textcolor{customgreen}{\gup 0.165}) & 0.608 (\textcolor{customgreen}{\gup 0.046}) & 0.617 (\textcolor{red}{\rdown 0.002}) & 1.092 (\textcolor{customgreen}{\gup 0.213}) & 0.613 (\textcolor{customgreen}{\gup 0.070}) & 0.583 (\textcolor{red}{\rdown 0.039}) & 0.576 (\textcolor{customgreen}{\gup 0.108}) & 0.587 (\textcolor{customgreen}{\gup 0.022}) & 0.510 (\textcolor{red}{\rdown 0.002}) \\
UIEC\textsuperscript{2}-Net \cite{WANG2021116250} & 23.347 (\textcolor{customgreen}{\gup 4.602}) & 0.860 (\textcolor{customgreen}{\gup 0.092}) & 0.822 (\textcolor{customgreen}{\gup 0.139}) & 0.610 (\textcolor{customgreen}{\gup 0.044}) & 0.631 (\textcolor{red}{\rdown 0.016}) & 0.893 (\textcolor{customgreen}{\gup 0.412}) & 0.619 (\textcolor{customgreen}{\gup 0.064}) & \textcolor{blue}{0.617} (\textcolor{red}{\rdown 0.073}) & 0.661 (\textcolor{customgreen}{\gup 0.023}) & 0.583 (\textcolor{customgreen}{\gup 0.026}) & 0.549 (\textcolor{red}{\rdown 0.041}) \\
UDAformer \cite{SHEN202377} & 25.350 (\textcolor{customgreen}{\gup 2.599}) & 0.921 (\textcolor{customgreen}{\gup 0.031}) & 0.759 (\textcolor{customgreen}{\gup 0.202}) & 0.596 (\textcolor{customgreen}{\gup 0.058}) & 0.606 (\textcolor{customgreen}{\gup 0.009}) & 0.949 (\textcolor{customgreen}{\gup 0.356}) & 0.609 (\textcolor{customgreen}{\gup 0.074}) & 0.570 (\textcolor{red}{\rdown 0.026}) & 0.543 (\textcolor{customgreen}{\gup 0.141}) & 0.561 (\textcolor{customgreen}{\gup 0.048}) & 0.502 (\textcolor{customgreen}{\gup 0.006}) \\
DM\_underwater \cite{10.1145/3581783.3612378} & 25.569 (\textcolor{customgreen}{\gup 2.380}) & 0.931 (\textcolor{customgreen}{\gup 0.021}) & 0.797 (\textcolor{customgreen}{\gup 0.164}) & 0.609 (\textcolor{customgreen}{\gup 0.045}) & 0.576 (\textcolor{customgreen}{\gup 0.039}) & 0.874 (\textcolor{customgreen}{\gup 0.431}) & 0.614 (\textcolor{customgreen}{\gup 0.069}) & 0.576 (\textcolor{red}{\rdown 0.032}) & 0.647 (\textcolor{customgreen}{\gup 0.037}) & 0.579 (\textcolor{customgreen}{\gup 0.030}) & 0.492 (\textcolor{customgreen}{\gup 0.016}) \\
\hline
CLIP-UIE (Baseline) \cite{liu2024underwaterimageenhancementdiffusion} & 25.412 (\textcolor{customgreen}{\gup 2.537}) & 0.936 (\textcolor{customgreen}{\gup 0.016}) & 0.981 (\textcolor{red}{\rdown 0.020}) & 0.619 (\textcolor{customgreen}{\gup 0.035}) & 0.624 (\textcolor{red}{\rdown 0.009}) & 1.257 (\textcolor{customgreen}{\gup 0.048}) & 0.645 (\textcolor{customgreen}{\gup 0.038}) & 0.544 (\textcolor{customgreen}{\gup 0.000}) & 0.754 (\textcolor{red}{\rdown 0.070}) & 0.588 (\textcolor{customgreen}{\gup 0.021}) & 0.497 (\textcolor{customgreen}{\gup 0.011}) \\
UDAN-CLIP (Ours) & \textcolor{blue}{27.949} & \textcolor{blue}{0.952} & 0.961 & \textcolor{blue}{0.654} & 0.615 & 1.305 & \textcolor{blue}{0.683} & 0.544 & 0.684 & \textcolor{blue}{0.609} & 0.508 \\
\bottomrule
\end{tabular}
}
\end{minipage}
\end{table*}
\end{landscape}
For the Color-Checker7 dataset, UDAN-CLIP demonstrates strong performance, particularly in enhancing color and contrast, as reflected by its leading UCIQE score of 0.683. This indicates the most effective color and contrast enhancement among all evaluated methods. The baseline model, CLIP-UIE, follows with a UCIQE of 0.645, while the traditional method UDCP comes next at 0.641. Other competing approaches fall behind, with values below 0.63, underscoring the substantial improvement offered by UDAN-CLIP in this aspect. In terms of UIQM, which captures overall perceptual quality in underwater imagery, UDAN-CLIP achieves the highest value among deep learning-based methods at 1.305, outperforming the baseline CLIP-UIE, which reports 1.257. Other learning-based methods, including TCTLNet, UIEC\textsuperscript{2}-Net, UDAformer, and DM\_underwater, all record UIQM scores below 1.1. This suggests that while UDAN-CLIP excels in perceptual quality, traditional methods such as UDCP still maintain an advantage in this specific metric. Regarding sharpness, as measured by CPBD, UIEC\textsuperscript{2}-Net achieves the highest score of 0.617, indicating the sharpest outputs with the least blur. UDCP follows with a score of 0.589, while UDAN-CLIP and CLIP-UIE report lower values of 0.544. This suggests a trade-off in UDAN-CLIP, where the model prioritizes perceptual and color quality over maximizing image sharpness, ultimately achieving a more balanced enhancement profile.

For the C60 dataset, which presents greater challenges due to lower contrast and severe color casts in the raw underwater images (as illustrated in Table \ref{tab:Quantitative_comp}), the competing methods generally demonstrate better generalization and more balanced performance across evaluation metrics. Among them, our proposed UDAN-CLIP model shows notable strength in both detail restoration and color correction. UDAN-CLIP achieves the highest UCIQE score of 0.609, outperforming all competing methods and indicating superior enhancement of color and contrast in this particularly challenging dataset. This highlights the model's robustness in handling images with complex distortions. In terms of perceptual sharpness, as measured by CPBD, UDAN-CLIP also performs well, achieving a higher score than the baseline CLIP-UIE (0.497) and DM-Underwater (0.492). While there is still room for improvement in sharpness, these results reflect UDAN-CLIP’s ability to maintain a strong balance between perceptual quality and structural enhancement under difficult underwater imaging conditions.
\begin{figure}[t!]
    \centering
    \includegraphics[width=\columnwidth]{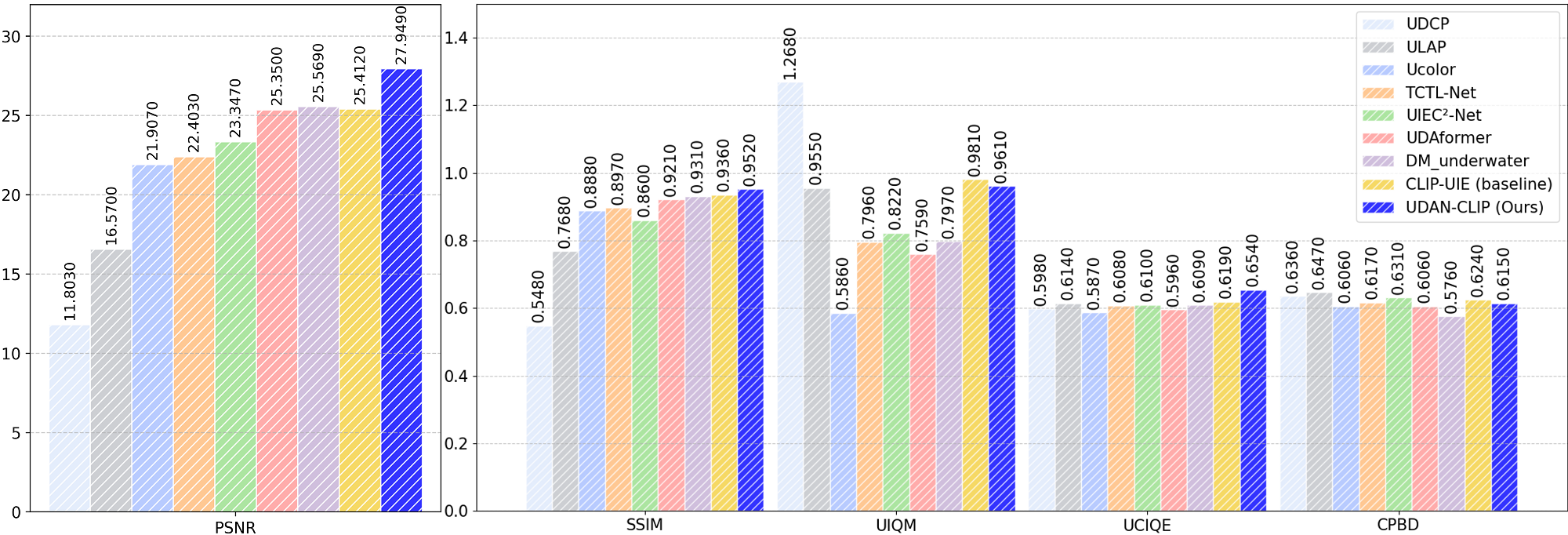}
    \caption{Performance metrics comparison of our proposed UDAN-CLIP model on the T200 dataset.}
    \label{fig:plot1}
\end{figure}

\begin{figure}[!b]
    \centering
    \includegraphics[width=\columnwidth]{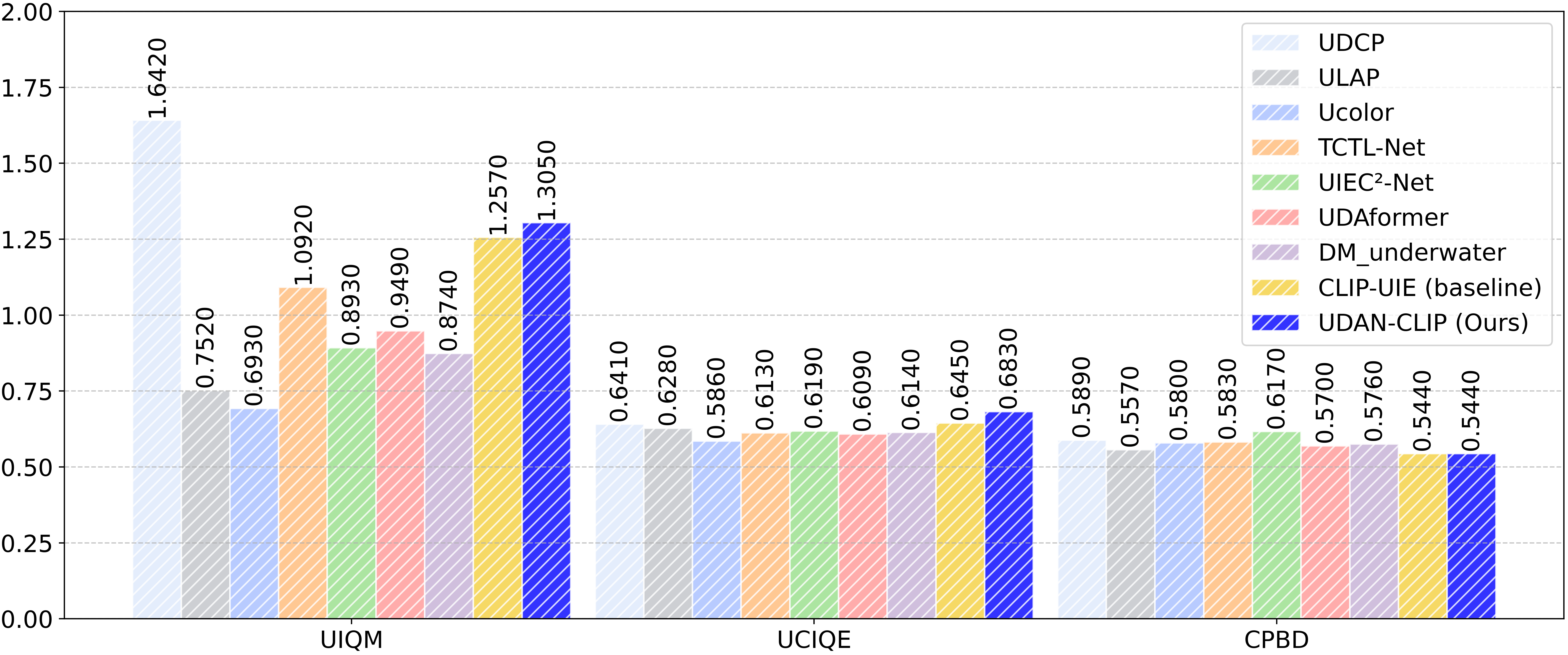}
    \caption{Performance metrics comparison of our proposed UDAN-CLIP model on the Color-Checker7 dataset.}
    \label{fig:plot2}
\end{figure}

\begin{figure}[!t]
    \centering
    \includegraphics[width=\columnwidth]{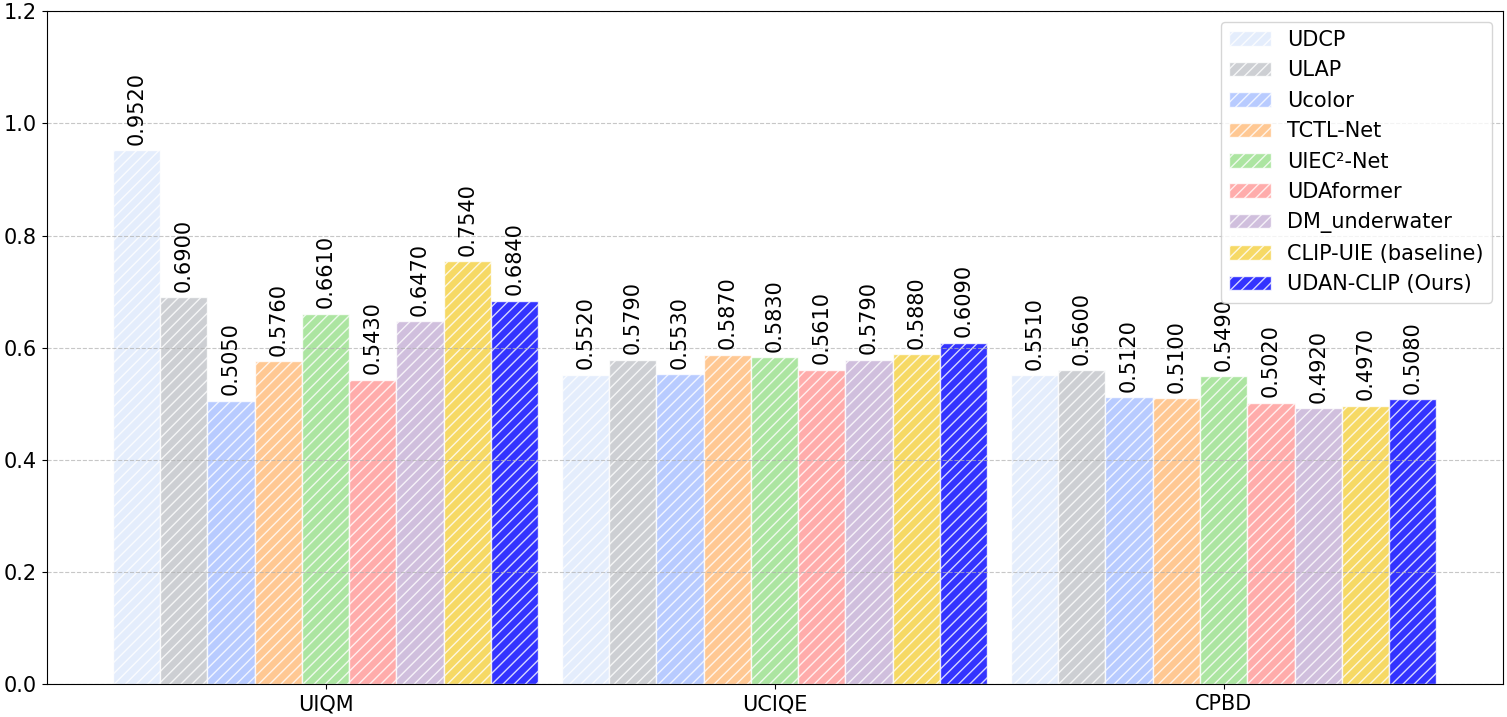}
    \caption{Performance metrics comparison of our proposed UDAN-CLIP model on the C60 dataset.}
    \label{fig:plot3}
\end{figure}

For a more comprehensive analysis of performance, bar plots are presented in  Figure \ref{fig:plot1}, Figure \ref{fig:plot2}, and Figure \ref{fig:plot3} for the T200, Color-Checker7, and C60 datasets, respectively. These visualizations offer a clear, metric-wise comparison of UDAN-CLIP against a range of state-of-the-art methods across each dataset. The bar plots highlight how UDAN-CLIP performs in terms of key evaluation metrics such as PSNR, SSIM, UIQM, UCIQE, and CPBD, providing detailed insights into its strengths in image reconstruction quality, structural preservation, perceptual enhancement, and sharpness.

\begin{figure}[b!]
    \centering
    \includegraphics[width=\columnwidth]{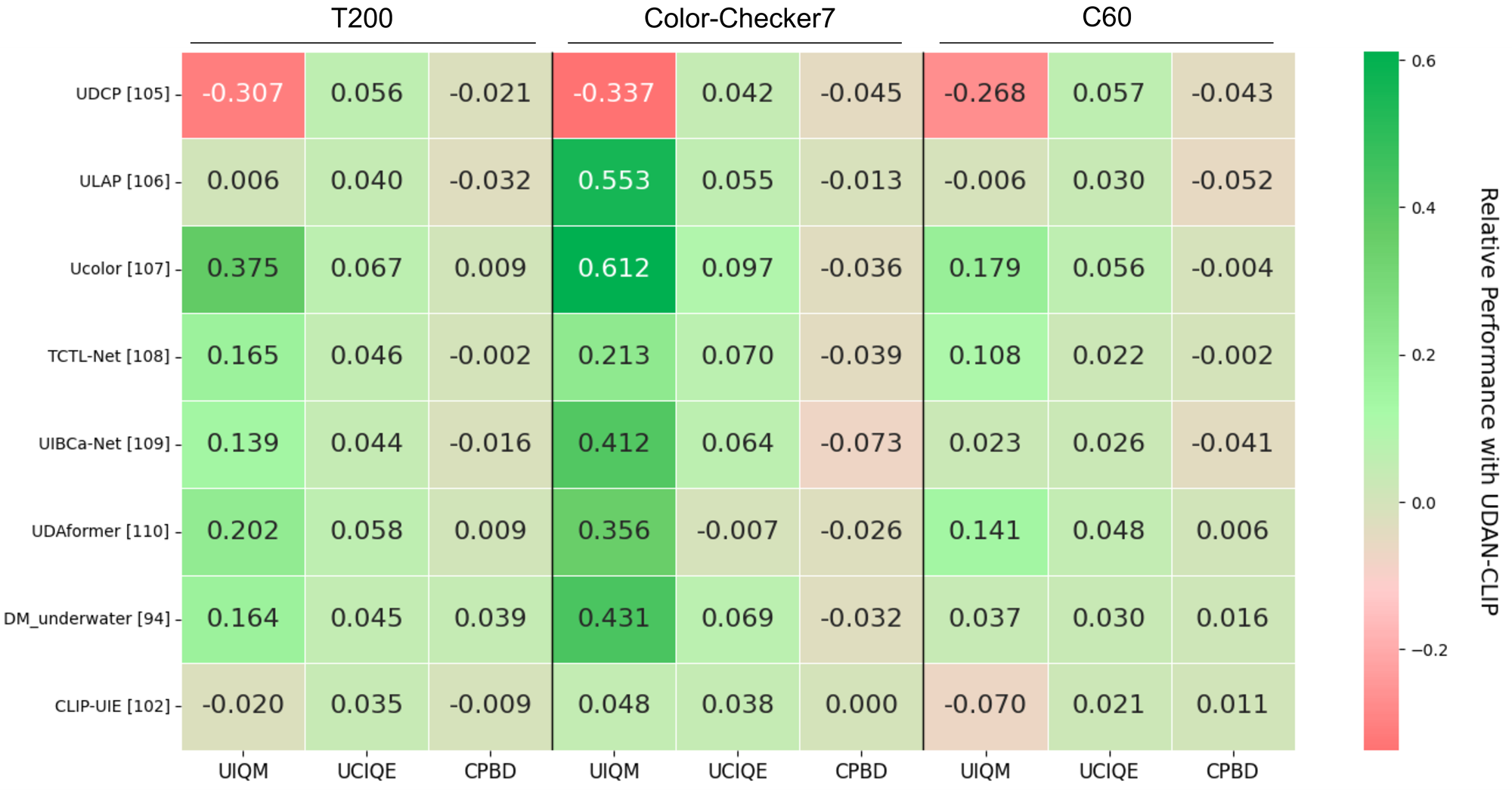}
    \caption{A heatmap illustrating the relative performance of UDAN-CLIP compared to baseline methods across the T200, Color-Checker7, and C60 datasets is presented using the UIQM, UCIQE, and CPBD metrics. In the heatmap, dark green indicates areas where UDAN-CLIP significantly outperforms other methods, light green represents neutral or moderate differences, and bright red highlights instances where competing methods surpass UDAN-CLIP. This visual comparison provides a clear overview of UDAN-CLIP’s strengths and areas for potential improvement across different datasets and evaluation criteria.} 
    \label{fig:heatmap}
\end{figure}

In addition to the bar plots, a heatmap is also shown in Figure \ref{fig:heatmap} to depict the relative performance of UDAN-CLIP compared to other leading methods across the three datasets. This heatmap uses color-coded cues: dark green indicating strong performance by UDAN-CLIP, light green showing moderate or neutral differences, and bright red highlighting cases where other methods outperform it. This comparative visualization helps identify trends across datasets and metrics, offering an intuitive summary of where UDAN-CLIP excels and where further improvements can be targeted. Together, these visual tools support a more detailed understanding of the performance profile of UDAN-CLIP in diverse underwater imaging conditions. 

\subsection{Qualitative Evaluations}

\begin{figure*}[b!]
    \centering
    \includegraphics[width=\textwidth]{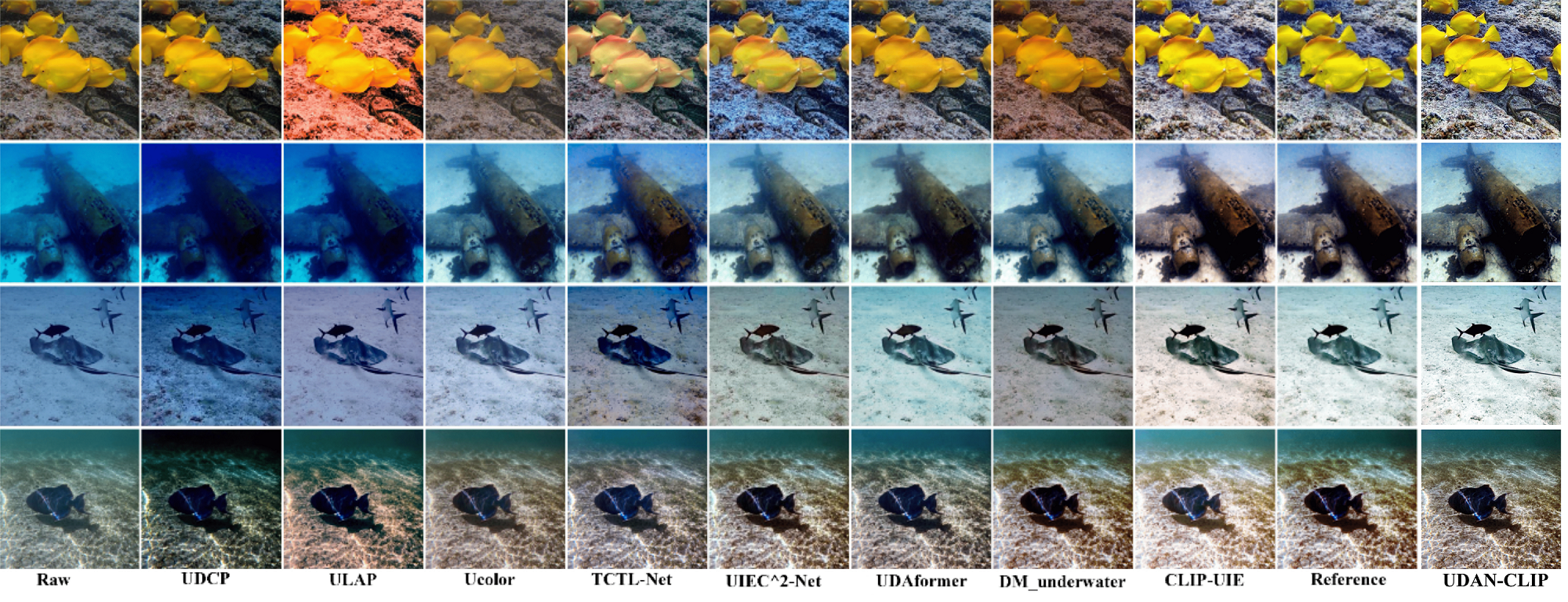}
    \caption{Qualitative comparison of the proposed method UDAN-CLIP (Ours) with traditional and deep learning-based approaches for underwater image enhancement on T200 datasets.}
    \label{fig:T200}
\end{figure*}

We present the visual outcomes of various comparison methods on the T200 dataset in Figure \ref{fig:T200}. Although traditional methods such as UDCP and ULAP often yield high scores on non-reference metrics, they typically introduce color distortions and suffer from noticeable loss of image detail, as evident in their results. In contrast, deep learning–based approaches benefit from the use of reference images during training, enabling them to generate visually appealing enhancements. However, this strategy can constrain the output to a manually defined reference domain, limiting generalization. As shown in the second and fourth rows of Figure \ref{fig:T200}, methods such as Ucolor \cite{9426457}, TCTL-Net \cite{10298280}, UIEC\textsuperscript{2}-Net \cite{WANG2021116250}, UDAformer \cite{SHEN202377}, DM\_underwater \cite{10.1145/3581783.3612378} produce highly similar visual results that often do not surpass the quality of the reference images.

In terms of color restoration and naturalness, UDAN-CLIP consistently delivers more vibrant and realistic colors compared to the baseline CLIP-UIE. As shown in Figure \ref{fig:T200}, in the first row (yellow fish), UDAN-CLIP preserves the bright yellow of the fish and the natural sand tones, while CLIP-UIE appears slightly duller. In the second row (submarine wreck), UDAN-CLIP more effectively reduces the blue/green cast, producing more lifelike water and object colors than CLIP-UIE.

In terms of color contrast and visual clarity, UDAN-CLIP consistently delivers images with higher contrast and more pronounced detail visibility. As shown in Figure \ref{fig:T200}, the third row featuring the stingray and fish illustrates this clearly, UDAN-CLIP produces a brighter, clearer background and more distinct outlines of the animals, whereas CLIP-UIE results appear flatter with less defined features. Similarly, in the fourth row showing the stingray on sand, UDAN-CLIP enhances both the brightness of the sand and the sharpness of the stingray’s outline, resulting in an image that more closely resembles the reference. Regarding color balance and white balance, UDAN-CLIP achieves a more accurate neutral tone, particularly in sandy and underwater regions, which appear more natural and less color-biased. In contrast, CLIP-UIE often retains a slight blue or green tint, detracting from the realism and true-to-life appearance of the images. Overall, as demonstrated in Figure~\ref{fig:T200} for the T200 dataset, UDAN-CLIP outputs are consistently closer to or better than the reference images, indicating superior enhancement of the scenes. The enhancements exhibit fewer visual artifacts and higher realism, demonstrating UDAN-CLIP’s superiority in both perceptual quality and scene fidelity.

\begin{figure*}[t]
    \centering
    \includegraphics[width=\textwidth]{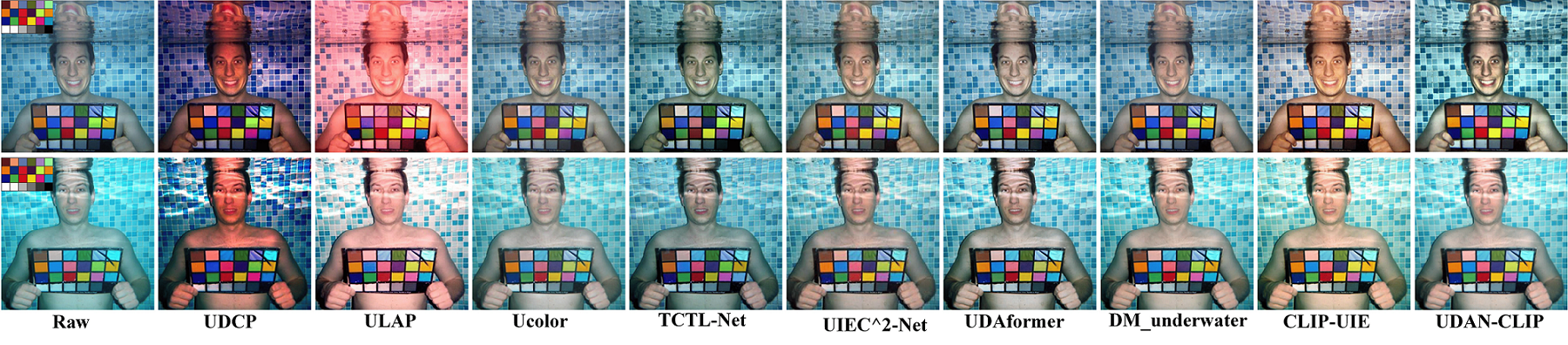}
    \caption{Qualitative comparison of our proposed method UDAN-CLIP with traditional and deep learning-based approaches for underwater image enhancement on the Color-Checker7 dataset.}
    \label{fig:color_checker7}
\end{figure*}

Similarly, a visual comparison between UDAN-CLIP and CLIP-UIE on the Color-Checker7 dataset is presented in Figure \ref{fig:color_checker7}. This dataset features underwater images with more severe color casts and reduced contrast, as clearly illustrated in the first row of the Figure \ref{fig:color_checker7}. Traditional methods such as UDCP and ULAP tend to introduce strong color distortions, resulting in unnatural skin tones and unrealistic color representation. Earlier deep learning–based approaches also struggle in this regard, often leaving residual blue or green tints in the enhanced images. These visual artifacts underscore the challenges posed by the dataset and highlight the importance of effective color correction and contrast enhancement, where UDAN-CLIP demonstrates clear advantages.

The color patches on the checkerboard in the UDAN-CLIP results as shown in Figure. \ref{fig:color_checker7} appear more vivid, distinct, and closely aligned with their true colors. The images show minimal color cast, with the checkerboard appearing more neutral and balanced. In contrast, the baseline CLIP-UIE results display slightly lower saturation and a mild blue or green tint, causing some colors to look washed out compared to UDAN-CLIP. In terms of skin tone, UDAN-CLIP produces more natural and lifelike results, effectively removing the typical underwater blue/green cast. Faces and body tones appear healthier and more realistic, with evenly distributed lighting. In comparison, CLIP-UIE retains a slight unnatural tint (cooler or greener) resulting in a less vibrant and slightly less realistic appearance. The background swimming pool tiles in the UDAN-CLIP outputs are also clearer, with accurate blue and white coloration and minimal distortion. The overall image looks crisp and natural. Meanwhile, in the CLIP-UIE results, the blues appear more muted and a faint color cast is still present, making the background less vibrant than in UDAN-CLIP. These observations demonstrate UDAN-CLIP’s superior ability to correct color casts, restore realistic tones, and enhance overall visual fidelity.

\begin{figure*}[t]
    \centering
    \includegraphics[width=\textwidth]{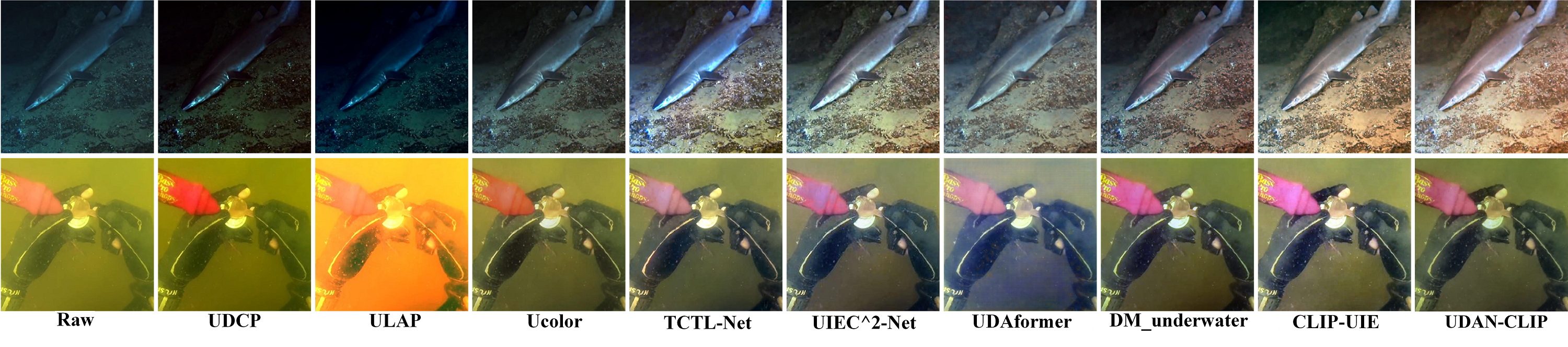}
    \caption{Qualitative comparison of our proposed method UDAN-CLIP with traditional and deep learning-based approaches for underwater image enhancement on the Color-Checker7 dataset.}
    \label{fig:C60}
\end{figure*}

A visual comparison between UDAN-CLIP and CLIP-UIE on the C60 dataset is shown in Figure \ref{fig:C60}. Overall, UDAN-CLIP provides a clearer, more focused, and visually appealing image. In the first row of the Figure \ref{fig:C60}, it can be noted that UDAN-CLIP produces a cooler, sharper, and higher-contrast image, emphasizing the shark features. However, the CLIP-UIE image suffers from blurriness, lower contrast, unnatural color shift, and less focus on the main subject. In the second row, it is evident that the UDAN-CLIP output demonstrates superior visual clarity and subject emphasis compared to the CLIP-UIE result. The UDAN-CLIP image features sharper outlines and more defined edges, particularly noticeable in the gloves and the object being held, which allows for easier identification of details such as the stitching on the gloves and the text on the pink object. The contrast is higher, making the hands and the object stand out distinctly from the background, which itself appears more uniform and less distracting. The color tone in UDAN-CLIP is cooler and more neutral, contributing to a more natural and realistic underwater appearance. In contrast, the CLIP-UIE image has a warmer, yellowish tint and appears slightly brighter but with a hazy overlay that reduces overall sharpness. The lower contrast in CLIP-UIE causes the gloves and object to blend more with the background, making fine details less noticeable and the text on the pink object harder to read. Additionally, the background in CLIP-UIE contains more visible noise and haze, which further detracts from the focus on the main subject.

\begin{figure*}[b]
    \centering
    \includegraphics[width=\textwidth]{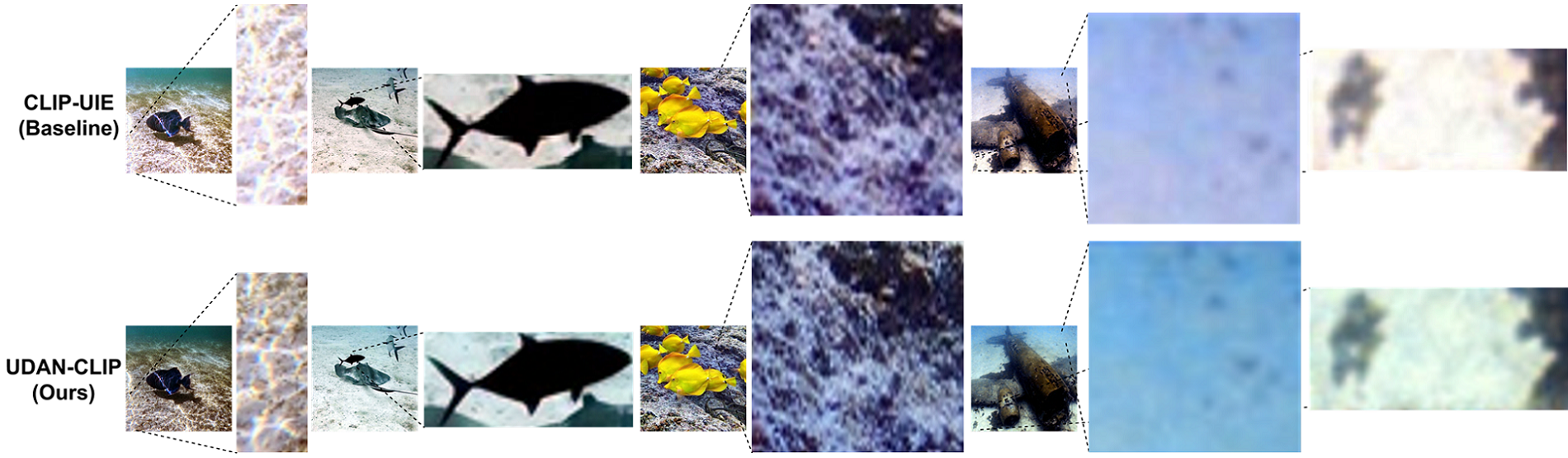}
    \caption{Region of Interest Qualitative comparison of the CLIP-UIE (Baseline) and UDAN-CLIP (Ours) on T200 dataset.}
    \label{fig:zoomedin_1}
\end{figure*}

\begin{figure*}[]
    \centering
    \includegraphics[width=\textwidth]{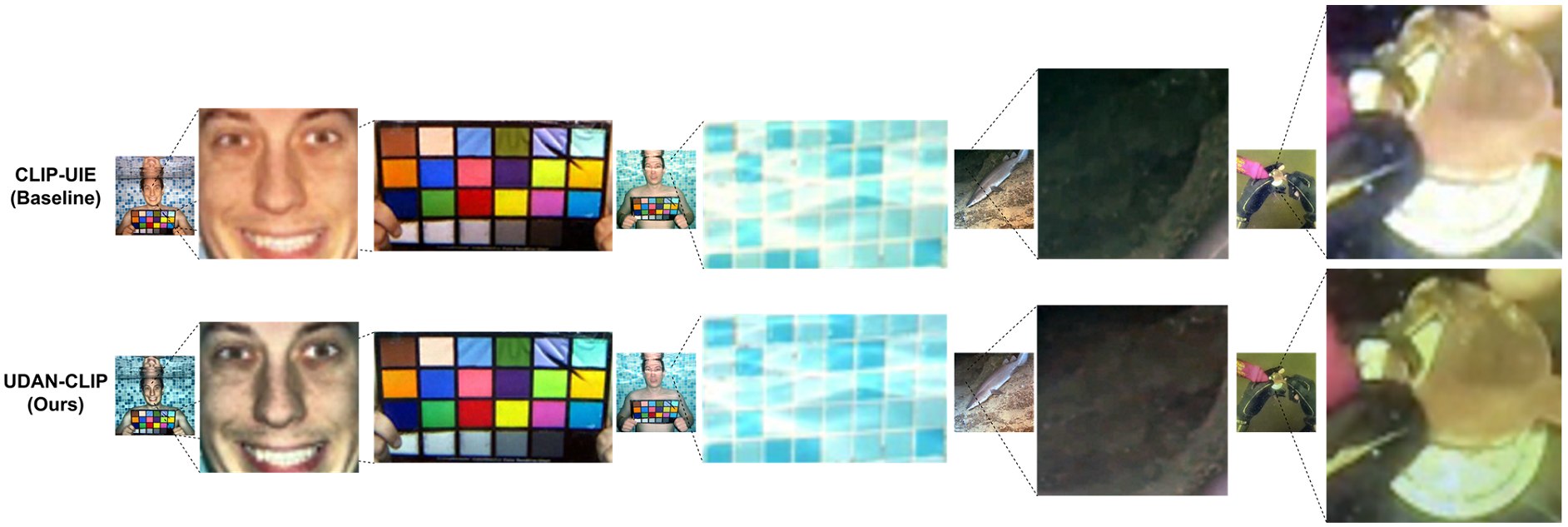}
    \caption{Region of Interest Qualitative comparison of the CLIP-UIE (Baseline) and UDAN-CLIP (Ours) on Color-Checker7 and C60 datasets.}
    \label{fig:zoomedin_2}
\end{figure*}

In Figure \ref{fig:zoomedin_1} and Figure \ref{fig:zoomedin_2}, we provide comparisons os zoomed-in regions of interest that offer a clear visual assessment of the enhancements made by UDAN-CLIP compared to the baseline CLIP-UIE across the T200, Color-Checker7, and C60 datasets. These focused views highlight UDAN-CLIP’s effectiveness in restoring fine details, enhancing contrast, and correcting color distortions. The results demonstrate that UDAN-CLIP consistently delivers more natural, sharper, and visually appealing outputs, especially in challenging underwater scenes where precise restoration is critical.

\subsection{Ablation Study}
We conduct a series of ablation studies using the T200 dataset to evaluate the effectiveness of the key components in the proposed UDAN-CLIP framework. These experiments help isolate and quantify the contribution of each module to the overall performance, providing insight into their individual and combined impact on underwater image enhancement.

\begin{table}[b!]
\centering
\caption{Quantitative results comparison of various network settings and optimization.}
\label{tab:Ablation}
\begin{tabular}{p{0.75\linewidth}|c|c}
\hline
\textbf{Ablation Component} & \textbf{PSNR} & \textbf{SSIM} \\
\hline
Baseline & 25.412 & 0.936 \\
Baseline + $\mathcal{L}_{\text{UDAN-CLIP}}$ & 26.322 & 0.941 \\
Baseline + $\mathcal{L}_{\text{UDAN-CLIP}}$ + Spatial Attention: UDAN-CLIP (Ours) & 27.949 & 0.952 \\
\hline
\end{tabular}
\end{table}

\subsubsection{Effectiveness of CLIP-Diffusion Loss $\mathcal{L}_{\text{UDAN-CLIP}}$}
We conducted experiments with the baseline and this novel loss function, and the results are presented in Table~\ref{tab:Ablation}. As shown in Table \ref{tab:Ablation}, we show the results of combining the baseline model with our proposed novel loss function $\mathcal{L}_{\text{UDAN-CLIP}}$. It can be seen that our model's capacity to bridge semantic gaps between generated and reference images is enhanced by the use of the novel CLIP-Diffusion loss. The loss aligns outputs with high-level concepts such as ``clear photo'' or ``normal light'' using pretrained vision-language embeddings, in contrast to standard pixel-level losses that prioritize low-frequency structural accuracy. Our model is able to suppress unrealistic artifacts (for instance, over-smoothed textures) by conditioning the diffusion process on these text-guided features. 

\subsubsection{Contribution of Spatial Attention in UDAN-CLIP}

Similarly, as observed in Table ~\ref{tab:Ablation}, we show the benefit of incorporating spatial-attention module into our model. It can be seen that our model's spatial attention module dynamically highlights areas rich in structural information, such as edges, textures, and high-frequency patterns, thereby improving feature extraction. The attention module generates soft masks to weight feature maps, allowing the model to focus on regions such as coral reefs in underwater images or facial contours in portraits. By preserving crucial information, this localization also enhances sharpness measurements. For instance, in turbid water scenes as shown in Figure \ref{fig:T200}, Figure \ref{fig:color_checker7}, and Figure \ref{fig:C60}, the attention module prioritizes sediment edges and the textures of marine life. This module effectively addresses challenges related to color distortion and contrast reduction, resulting in improved quality of underwater images. The results in Table~\ref{tab:Ablation} concludes that the integration of the CLIP-diffusion loss and spatial attention module progressively improves performance, with the full UDAN-CLIP model achieving the highest PSNR and SSIM, confirming the effectiveness of all components.

\section{Conclusion}

We proposed UDAN-CLIP, a real-world underwater image enhancement framework that combines a diffusion-based image-to-image model with a customized classifier based on vision-language model, and spatial attention module. The diffusion model captures the mapping from underwater to natural in-air images, while the classifier guides semantic alignment and stability during fine-tuning. The spatial attention module enables targeted enhancement of degraded regions. Extensive evaluations show that UDAN-CLIP effectively reduces common underwater distortions and outperforms existing methods in both visual quality and quantitative benchmarks for underwater image enhancement.


\bibliographystyle{plainnat}
\bibliography{cas-refs}

\end{document}